\documentclass{article}

\usepackage{iclr2027_conference,times}
\usepackage{amsmath,amssymb}
\usepackage{booktabs}
\usepackage{graphicx}
\usepackage{hyperref}
\usepackage{url}
\usepackage{booktabs}
\usepackage{multirow}
\usepackage{graphicx}
\usepackage[table]{xcolor}
\usepackage{pifont}
\usepackage{kotex} % 한국어 원고인 경우
\usepackage{adjustbox}
\usepackage{wrapfig}
\usepackage{float}
\hypersetup{hidelinks}
\newcommand{\cmark}{\ding{51}}
\newcommand{\xmark}{\ding{55}}

\usepackage{pgfplots}
\usepgfplotslibrary{groupplots}
\pgfplotsset{compat=1.18}

\title{
LEARN-TS: LLM-Enhanced Alignment and 

Reconstruction with Normality Guidance
for Multivariate Time-Series Anomaly Detection
}

\author{
Jahyeob Koo \quad
Kio Yun \quad
Byoungmo Koo \quad
Jun-Geol Baek \\
Department of Industrial and Management Engineering, Korea University \\
Seoul, Republic of Korea \\
\texttt{\{rnwkguq1506, ykio, kbm970709, jungeol\}@korea.ac.kr}
}
\iclrfinalcopy

\begin{document}

\maketitle
\lhead{}

\begin{abstract}
Reconstruction errors in multivariate time-series anomaly detection
may not reliably distinguish abnormal behavior from benign deviations.
Language-derived semantics offer complementary context, but existing
multimodal approaches may rely on time associated paired textual information
that is difficult to obtain consistently and is not provided by standard
multivariate time-series anomaly detection benchmarks.
This setting poses two challenges:
(1) conditioning masked reconstruction on window-specific semantics without
exposing exact numerical targets or anomaly specific cues, and
(2) using a window-independent concept of normality as a complementary
semantic reference rather than an independent anomaly detector. We propose LLM-Enhanced Alignment and Reconstruction with Normality Guidance
for Time Series (LEARN-TS), which uses a frozen language model to construct
two role separated semantic representations without requiring temporally paired external text.
Window-specific observation semantics encode temporal and
cross-variable context without exact numerical values to guide
channel-shared patch-masked reconstruction.
A fixed, dataset-agnostic normality prompt provides a window-independent
semantic reference for aligning normal representations and estimating
normality discrepancy.
At inference, masking each temporal patch once yields timestamp level
reconstruction evidence, conditionally modulated by discrepancy from a
separate unmasked view. Across four benchmarks, LEARN-TS achieves the highest mean
performance in 13 of 16 dataset--metric comparisons.
Controlled ablations examine observation conditioning, joint
normality alignment and scoring, and reference content, showing
dataset-dependent ranking benefits and modest average gains
from semantic over random references.
\end{abstract}

\section{Introduction}
\label{sec:introduction}

Multivariate time-series anomaly detection (MTSAD) is essential for monitoring
industrial processes, computing infrastructures, application services, and
spacecraft systems.
Because anomalous events are rare and costly to annotate, MTSAD is commonly
formulated under unsupervised or one-class settings, where normal operation
patterns are learned and deviations are identified from reconstruction,
prediction, or representation based evidence.
A fundamental difficulty, however, is that numerical deviation alone does not
fully characterize whether an observation is behaviorally abnormal.
Expressive models may reconstruct anomalous inputs accurately, while benign
regime changes or distribution shifts may produce large residuals.
Consequently, anomaly evidence derived only from numerical discrepancies can
remain ambiguous without context about what constitutes coherent normal
behavior, motivating the use of complementary semantic information.

\begin{figure*}[t]
    \centering
    \includegraphics[width=1.0\textwidth]{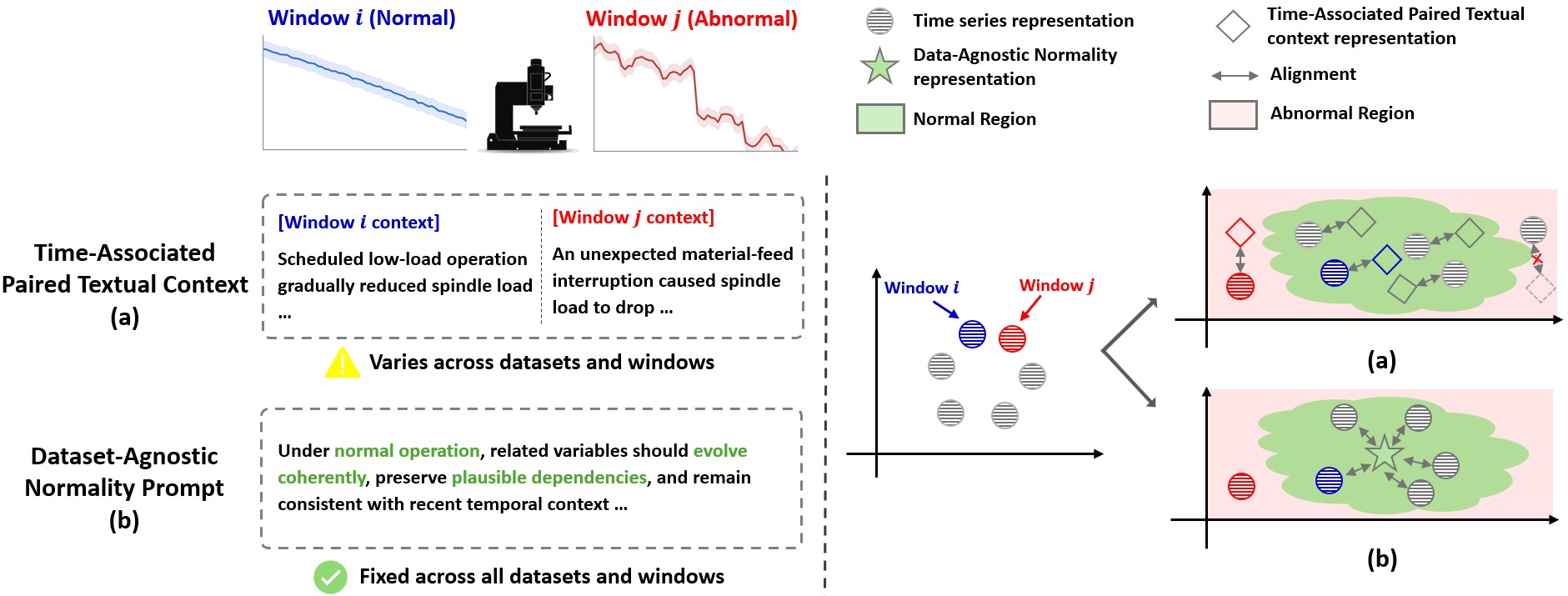}
    \caption{
(a) Time-associated paired external textual context may vary across datasets and
monitored intervals, providing context-dependent semantic information.
(b) LEARN-TS instead derives a shared, window-independent normality reference
from a fixed, dataset-agnostic prompt applied unchanged across all datasets
and windows.
    }
    \label{fig:motivation}
\end{figure*}

Recent language model and multimodal time-series methods have begun to
introduce such semantic information into time-series analysis.
In multimodal anomaly detection, semantic context may be provided through
textual descriptions, reports, logs, or other information associated with
particular time periods or events.
This form of \emph{time associated paired textual context} can provide rich
behavioral cues, but it is difficult to assume in many industrial monitoring
settings.
Operational logs and reports are often sparse, irregularly recorded, written
at a different temporal granularity from sensor streams, or unavailable for
large portions of the monitored timeline.
Moreover, standard MTSAD benchmarks such as SWaT, SMD, PSM, and MSL do not
provide paired textual observations for individual windows.
Requiring such context therefore changes the information available to the
detector and limits applicability when comprehensive temporal pairing cannot
be obtained.
This raises a practical question: \textbf{can language-derived semantics
provide useful guidance without requiring temporally paired
external text?}

We address this question by assigning distinct roles to two complementary
semantic representations.
First, \emph{window-specific observation semantics} summarize coarse temporal
and cross-variable characteristics of each input window without including
exact numerical values or explicit anomaly labels.
Second, a \emph{window-independent normality reference} is derived from a fixed,
dataset-agnostic prompt shared across all datasets and windows.
As illustrated in Figure~\ref{fig:motivation}, panel (a) assumes
time-associated paired external text, whereas panel (b) uses the fixed normality
prompt without requiring interval-specific external textual context.

Building on this formulation, we propose
\textbf{LLM-Enhanced Alignment and Reconstruction with Normality Guidance for
Time Series (LEARN-TS)}.
It uses a frozen language model to construct two complementary semantic
representations: window-specific observation semantics that guide numerical
reconstruction, and a shared semantic normality reference that supports
normality-aware representation learning and anomaly scoring.
Across SWaT, SMD, PSM, and MSL, LEARN-TS achieves the best result in 13 of 16
dataset--metric comparisons, demonstrating strong performance across both
ranking-based and event-level evaluation criteria.
Controlled analyses examine cross-modal fusion, observation--window
correspondence, the roles of normality alignment and scoring,
and the value of structured normality semantics relative to a random
semantic-free reference.

Our contributions are summarized as follows:

\begin{itemize}
\item We introduce a semantic guidance framework for MTSAD that
assigns distinct roles to window-specific observation descriptions
and a shared normality prompt, without requiring temporally paired
external text.

\item We develop observation-conditioned patch-masked reconstruction
and normality-guided representation alignment, combining exhaustive
patch-wise reconstruction evidence with discrepancy-based score
modulation from a separate unmasked view.

\item Through five-seed benchmark evaluations and controlled ablations
on four datasets, we examine observation conditioning, joint
alignment and scoring, and reference content. The results show dataset-dependent
ranking benefits and modest average gains from semantic over
random references.
\end{itemize}

% 시간적으로 짝지어진 외부 텍스트를 요구하지 않으면서, window별 observation 설명과 공유 normality prompt에 서로 다른 역할을 부여하는 MTSAD semantic guidance 프레임워크를 제안합니다.
% Observation 기반 patch-masked reconstruction과 정상성 기반 표현 정렬을 개발하고, 모든 patch를 차례로 masking하여 얻은 복원 근거에 별도 unmasked view의 discrepancy를 이용한 점수 조절을 결합합니다.
% 네 벤치마크의 five-seed 실험과 대조 ablation을 통해 observation conditioning, 정렬·scoring 결합, reference 내용의 효과를 분석합니다. 결과는 데이터셋별 ranking 이점과 random 대비 semantic reference의 소폭 평균 개선을 보여줍니다.

% \begin{itemize}

% \item We formulate a semantic MTSAD setting that does not require
% time associated paired external textual context, and introduce a role separated
% framework combining window-specific observation semantics with a
% window-independent, dataset-agnostic normality reference.

% \item We propose semantic-context-conditioned reconstruction in which
% exact value free observation semantics provide complementary temporal and
% cross variable context for numerical anomaly evidence.

% \item We introduce normality-guided representation alignment and
% discrepancy-based score modulation using a window-independent
% semantic reference derived from a fixed, dataset-agnostic
% normality prompt.

% \end{itemize}

\section{Related Work}
\label{sec:related_work}

\subsection{Reconstruction- and Normality-Based Time-Series Anomaly Detection}

Unsupervised MTSAD typically models normal behavior through reconstruction
or prediction.
Representative methods use stochastic latent dynamics
\citep{su2019omnianomaly}, adversarially trained autoencoders
\citep{audibert2020usad}, or association discrepancy alongside reconstruction
evidence~\citep{xu2022anomalytransformer}.
Recent work addresses distribution drift, reconstruction
over generalization, and cross-dataset generalization
\citep{wang2023drift,sun2025igad,shentu2025dada}, while memory-based and
one-class methods encode normal patterns using memory items, prototypes,
latent centers, or hierarchical hyperspheres
\citep{song2023memto,li2023puad,ruff2018deepsvdd,shen2020thoc}.
These methods derive normality primarily from numerical data; LEARN-TS adds
a language-derived semantic normality reference.

\subsection{Language Models for Time-Series Analysis}

Pretrained language models have been adapted as general time-series
backbones in GPT4TS~\citep{zhou2023gpt4ts}, through numerical patch
reprogramming in Time-LLM~\citep{jin2024timellm}, and with multiscale
language-derived context in LLM-Mixer~\citep{kowsher2025llmmixer}.
However, removing or replacing the language model may preserve forecasting
performance in some settings~\citep{tan2024language}.
LEARN-TS instead uses a frozen language model to encode explicitly
constructed semantic descriptions rather than primarily as a numerical
backbone.

\subsection{Multimodal and Semantic Alignment for Time-Series Anomaly Detection}

Multimodal time-series methods augment numerical observations with textual or
other semantic information and learn correspondences across modalities.
CALF applies cross-modal alignment and fine-tuning to time-series forecasting
\citep{liu2025calf}.
For anomaly detection, MindTS combines endogenous time-series descriptions
with externally collected background text through fine-grained semantic
alignment, condensed interaction, and cross-modal reconstruction
\citep{hu2026mindts}.
MindTS uses external background text and supports loose or sparse
window-level matching rather than requiring strict temporal alignment.

LEARN-TS does not require a corpus of external text associated
with monitored intervals.
Instead of associating external textual context with individual windows,
LEARN-TS derives observation semantics from each numerical input and uses a
window-independent normality reference constructed from a fixed,
dataset-agnostic prompt.
The two semantic representations are assigned distinct roles in
reconstruction conditioning, normality alignment, and anomaly scoring.

\section{Methodology}
\label{sec:methodology}

Let
\(\mathbf{X}=(\mathbf{x}_1,\ldots,\mathbf{x}_T)
\in\mathbb{R}^{T\times C}\)
denote a multivariate input sequence with \(T\) timestamps and \(C\)
variables, where \(\mathbf{x}_t\in\mathbb{R}^{C}\).
LEARN-TS operates on \(N\) sliding windows
\(\mathbf{X}_i\in\mathbb{R}^{L\times C}\),
\(i\in\{1,\ldots,N\}\), where \(L\) denotes the window length and
\(N\) the total number of windows extracted from the input sequence.
At inference, it produces a timestamp-level anomaly score \(S_t\) and
classifies timestamp \(t\) as anomalous when
\(\widehat{y}_t=\mathbb{I}[S_t>\tau]\),
where \(\tau\) is selected according to the evaluation protocol
in Section~\ref{sec:experimental_setting}.

\begin{figure*}[t]
    \centering
    \includegraphics[
        width=1\textwidth,
    ]{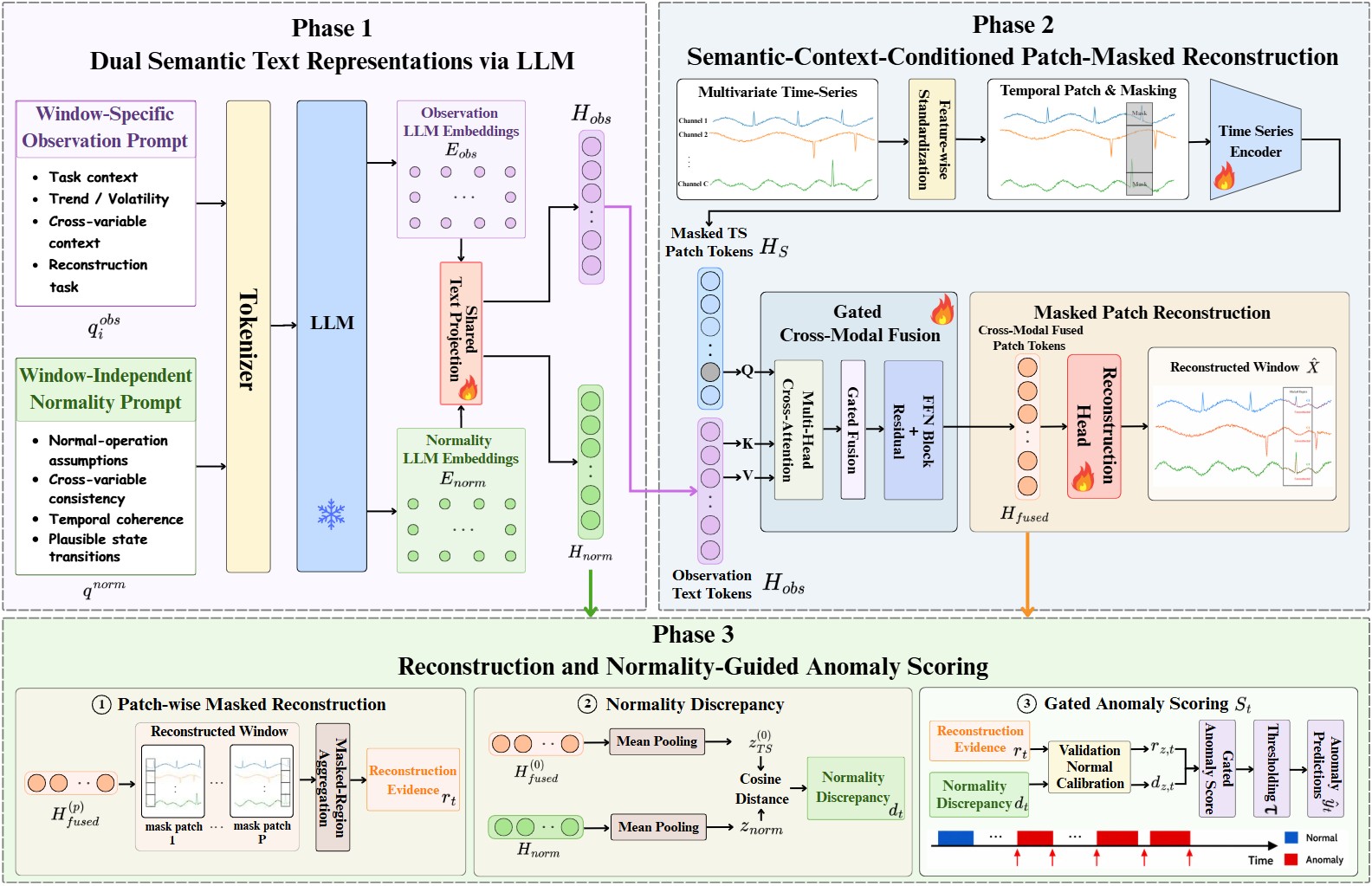}
    \caption{Overview of the three-phase LEARN-TS framework.}
    \label{fig:framework}
\end{figure*}

% =========================================================
% 3.1 Overall Framework
% =========================================================
\subsection{Overall Framework}
\label{sec:overall_framework}

As illustrated in Figure~\ref{fig:framework}, LEARN-TS comprises three
stages.
First, a frozen language model encodes a window-specific observation prompt
and a fixed, dataset-agnostic normality prompt shared unchanged across all
datasets and windows.
Second, the observation representation conditions channel shared
patch masked reconstruction through gated cross-modal fusion, while fused
representations of normal training windows are aligned with the normality
representation.
Finally, inference combines reconstruction evidence obtained by masking each
temporal patch once with normality discrepancy computed from a separate
unmasked pass.
The calibrated discrepancy conditionally modulates reconstruction evidence
to produce timestamp-level anomaly scores.

% =========================================================
% 3.2 Dual Semantic Representations
% =========================================================
\subsection{Dual Semantic Text Representations via LLM}
\label{sec:dual_semantic_representations}

\paragraph{Window-specific observation representation.}
For each window \(\mathbf{X}_i\), the observation prompt
\(q_i^{\mathrm{obs}}\) combines deterministic descriptors of the
fully observed window with fixed dataset-level system descriptions,
sensor groups, and qualitative operational rules.
Descriptors summarize trends, volatility, and temporal and
cross-variable relationships without exposing exact measurements,
timestamps, change-point locations, local deviation magnitudes,
anomaly labels, or explicit anomaly judgments.
Only the normality prompt is dataset-agnostic.
Construction details appear in Appendix~\ref{app:observation_prompt}.

\paragraph{Window-independent normality representation.}
To construct a window-independent semantic reference, we define a fixed,
dataset-agnostic normality prompt \(q^{\mathrm{norm}}\) describing broad
properties of normal multivariate temporal behavior, including temporal
coherence, consistent changes among related variables, plausible state
transitions, and context supported recovery.
The prompt contains no dataset-specific sensor identities, anomaly labels,
or test related information and is shared unchanged across all datasets and
input windows.
It therefore induces a window-independent semantic notion of normality rather
than context associated with a particular observed interval.

\paragraph{Frozen text encoding and shared projection.}
A frozen language model $E_{\mathrm{text}}$ encodes both prompts,
and a shared trainable projection $\phi(\cdot)$ maps their token
embeddings to the time-series model dimension:
\begin{equation}
\mathbf{H}_{\mathrm{obs},i}
=
\phi\!\left(E_{\mathrm{text}}(q_i^{\mathrm{obs}})\right),
\qquad
\mathbf{H}_{\mathrm{norm}}
=
\phi\!\left(E_{\mathrm{text}}(q^{\mathrm{norm}})\right).
\label{eq:semantic_representations}
\end{equation}
Here, $\mathbf{H}_{\mathrm{obs},i}\in\mathbb{R}^{T_o\times d}$
and $\mathbf{H}_{\mathrm{norm}}\in\mathbb{R}^{T_n\times d}$,
where $T_o$ and $T_n$ denote the respective token lengths and
\(d\) the projected time-series model dimension.
The frozen embeddings are precomputed and cached, while the shared
projection adapts them to the learned time-series space.
For a given backbone, the normality prompt and its frozen embedding
are fixed across datasets and windows; the projected normality
reference is learned per model and shared across its windows.
Model configurations and prompt construction details are provided
in Appendices~\ref{app:implementation_details}
and~\ref{app:prompt_templates}, respectively.

% =========================================================
% 3.3 Semantic-Conditioned Reconstruction
% =========================================================
\subsection{Semantic-Context-Conditioned Patch Masked Reconstruction}
\label{sec:semantic_conditioned_reconstruction}

LEARN-TS employs an asymmetric text-conditioned reconstruction formulation.
The numerical branch masks the same temporal region across all variables to
suppress direct numerical copying, whereas the semantic branch retains
coarse context derived from the fully observed window.
Numerical masking is therefore intended to suppress direct identity mapping
rather than to formulate information free missing value imputation.

\paragraph{Channel shared temporal patch masking.}
Let
\(\overline{\mathbf{X}}_i\in\mathbb{R}^{L\times C}\)
denote a feature-wise standardized input window.
We divide it into
\(P=1+\lfloor(L-\ell)/s\rfloor\) temporal patches of length \(\ell\) and
stride \(s\), indexed by \(p\in\{1,\ldots,P\}\).
Each patch contains all \(C\) variables over its temporal interval and is
flattened and linearly projected with a positional embedding to form
\(\mathbf{z}_{i,p}\in\mathbb{R}^{d}\).

During training, one temporal patch \(m_i\) is sampled uniformly and replaced
with a trainable mask token:
\begin{equation}
m_i\sim\operatorname{Uniform}\{1,\ldots,P\},
\qquad
\widetilde{\mathbf z}_{i,p}^{(m_i)}
=
\begin{cases}
\mathbf z^{\mathrm{mask}}, & p=m_i,\\
\mathbf z_{i,p}, & p\neq m_i.
\end{cases}
\label{eq:stochastic_patch_masking}
\end{equation}
Because each patch token jointly represents all variables, this operation
masks the same temporal region across every variable and prevents
cross variable copying at the held out timestamps.
A time-series encoder maps the masked token sequence to numerical patch
representations
\(\mathbf{H}_{S,i}^{(m_i)}\in\mathbb{R}^{P\times d}\). Here, \(S\) denotes the time-series branch, \(i\) indexes the input window,
and \(P\) is the number of temporal patches.

\paragraph{Selective semantic conditioning.}
Starting from
\(\mathbf{U}_i^{[0]}=\mathbf{H}_{S,i}^{(m_i)}\), where \(\mathbf{U}^{[0]}_i\) denotes the initial numerical representation, each fusion block obtains
\(\mathbf{A}_i^{[k]}\) through multi-head cross-attention using
\(\mathbf{U}_i^{[k-1]}\) as queries and
\(\mathbf{H}_{\mathrm{obs},i}\) as keys and values.
Because the relevance of semantic context may vary across patches,
\(\mathbf{A}_i^{[k]}\) is selectively integrated through gated residual
updates:
\begin{equation}
\begin{aligned}
\mathbf G_i^{[k]}
&=
\sigma\!\left(
\operatorname{MLP}
([\mathbf U_i^{[k-1]};\mathbf A_i^{[k]}])
\right),\\
\mathbf V_i^{[k]}
&=
\mathbf G_i^{[k]}\odot\mathbf A_i^{[k]}
+
(\mathbf 1-\mathbf G_i^{[k]})\odot\mathbf U_i^{[k-1]},\\
\mathbf R_i^{[k]}
&=
\operatorname{LN}_1
(\mathbf U_i^{[k-1]}+\mathbf V_i^{[k]}),
\qquad
\mathbf U_i^{[k]}
=
\operatorname{LN}_2
(\mathbf R_i^{[k]}+\operatorname{FFN}(\mathbf R_i^{[k]})).
\end{aligned}
\label{eq:gated_residual_fusion}
\end{equation}
Here, \(\mathbf{A}_i^{[k]}\) denotes the observation-conditioned
cross-attention output of the \(k\)-th fusion block,
\([\cdot;\cdot]\) denotes concatenation along the feature dimension,
\(\sigma\) is the sigmoid function,
\(\odot\) denotes element-wise multiplication, and
\(\operatorname{LN}_1\) and \(\operatorname{LN}_2\) denote the two
LayerNorm operations.
After \(K\) fusion blocks, the final fused representation is \(\mathbf H_{\mathrm{fused},i}^{(m_i)}=\mathbf U_i^{[K]}\), where \(K\) denotes the number of fusion blocks.  

\paragraph{Patch-wise reconstruction.}
A lightweight LayerNorm--linear reconstruction head maps each fused token
back to its standardized multivariate patch.
The reconstructed patches are combined to form
\(\widehat{\mathbf{X}}_i^{(m_i)}
\in\mathbb{R}^{L\times C}\).
The same time-series encoder, fusion blocks, and reconstruction head are
reused under the patch mask conditions employed at inference.
Model and training configurations are provided in
Appendix~\ref{app:implementation_details}.

% =========================================================
% 3.4 Optimization and Scoring
% =========================================================
\subsection{Joint Optimization and Normality-Guided Scoring}
\label{sec:optimization_and_scoring}

\paragraph{Training objective.}
LEARN-TS jointly optimizes masked-patch reconstruction, auxiliary full-window
reconstruction, and semantic normality alignment.
Let \(\Omega_{\mathrm{mask}}\) denote the masked timestamp--variable
positions in a training batch and let \(\Omega_{\mathrm{full}}\) denote all
positions.
For \(a\in\{\mathrm{mask},\mathrm{full}\}\), the reconstruction losses are
\begin{equation}
\mathcal L_a
=
\frac{1}{|\Omega_a|}
\sum_{(i,t,c)\in\Omega_a}
\left(
\widehat X_{i,t,c}^{(m_i)}
-
\overline X_{i,t,c}
\right)^2.
\label{eq:reconstruction_losses}
\end{equation}
Here, \(c\in\{1,\ldots,C\}\) indexes the variables. For a token sequence \(\mathbf H\), define
\(\rho(\mathbf H)
=\operatorname{norm}(\operatorname{MeanPool}(\mathbf H))\), with
\(\operatorname{norm}(\mathbf v)
=\mathbf v/(\lVert\mathbf v\rVert_2+\epsilon)\),
where \(\epsilon>0\) is a small constant for numerical stability.

The masked fused representation and normality representation are summarized
as
\begin{equation}
\mathbf z_{\mathrm{TS},i}^{(m_i)}
=
\rho(\mathbf H_{\mathrm{fused},i}^{(m_i)}),
\qquad
\mathbf z_{\mathrm{norm}}
=
\rho(\mathbf H_{\mathrm{norm}}).
\label{eq:pooled_representations}
\end{equation}
Because the training windows represent normal operation, their masked fused
representations are encouraged to remain close to the shared semantic
normality reference, where \(\mathcal B\) denotes the current mini-batch:
\begin{equation}
\mathcal L_{\mathrm{norm}}
=
\frac{1}{|\mathcal B|}
\sum_{i\in\mathcal B}
\left[
1-
\cos\!\left(
\mathbf z_{\mathrm{TS},i}^{(m_i)},
\mathbf z_{\mathrm{norm}}
\right)
\right].
\label{eq:normality_alignment}
\end{equation}
Applying normality alignment to randomly masked views encourages robustness
to stochastic patch corruption.
The complete training objective is
\begin{equation}
\mathcal L
=
\mathcal L_{\mathrm{mask}}
+
\lambda_{\mathrm{full}}\mathcal L_{\mathrm{full}}
+
\lambda_{\mathrm{norm}}\mathcal L_{\mathrm{norm}}.
\label{eq:total_objective}
\end{equation} Here, $\lambda_{\mathrm{full}}$ and $\lambda_{\mathrm{norm}}$ weight the
full-window reconstruction and normality-alignment losses, respectively.
Although no direct alignment loss is imposed on
$\mathbf{H}_{\mathrm{obs},i}$, the observation projection and fusion modules
are jointly optimized through the reconstruction and normality-alignment
objectives. The language model remains frozen.

\paragraph{Exhaustive patch-wise reconstruction evidence.}
At inference, LEARN-TS masks each temporal patch once.
Let \(\widehat{\mathbf X}_i^{(p)}\) denote the reconstruction obtained when
only patch \(p\) is masked.
Under the non-overlapping patch configuration used in our experiments, let
\(p(t)\) denote the unique patch containing timestamp \(t\).
The reconstruction evidence is
\begin{equation}
r_{i,t}
=
\mathcal{G}_{\mathcal D}
\left(
\left\{
\left(
\widehat{X}_{i,t,c}^{(p(t))}
-
\overline{X}_{i,t,c}
\right)^2
\right\}_{c=1}^{C}
\right),
\label{eq:reconstruction_evidence}
\end{equation}
where \(\mathcal{G}_{\mathcal D}\) is the dataset-specific channel error
aggregation operator defined in
Appendix~\ref{app:implementation_details}.
Thus, \(r_{i,t}\) uses only the pass in which \(p(t)\) is masked;
contributions from overlapping sliding windows are aggregated on the original
timeline to obtain \(r_t\).

\paragraph{Unmasked normality discrepancy.}
A separate unmasked inference pass produces
\(\mathbf H_{\mathrm{fused},i}^{(0)}\), from which the window-level
normality discrepancy is computed as
\begin{equation}
\mathbf z_{\mathrm{TS},i}^{(0)}
=
\rho(\mathbf H_{\mathrm{fused},i}^{(0)}),
\qquad
d_i
=
1-
\cos\!\left(
\mathbf z_{\mathrm{TS},i}^{(0)},
\mathbf z_{\mathrm{norm}}
\right).
\label{eq:normality_discrepancy}
\end{equation}
Masked-view alignment during training encourages robustness to patch
corruption, whereas the unmasked inference view makes \(d_i\) independent of
a particular mask position.
Each \(d_i\) is assigned to the timestamps covered by its window, and
overlapping contributions are aggregated to obtain \(d_t\).
The dataset-specific aggregation rules are reported in
Table~\ref{tab:score_aggregation}.

\paragraph{Normality-gated anomaly score.}
Because reconstruction evidence and normality discrepancy have different
scales, we independently calibrate them using normal validation score
distributions, obtaining \(r_{z,t}\) and \(d_{z,t}\), which denote the
calibrated reconstruction evidence and calibrated normality discrepancy,
respectively.
The calibration procedure is provided in
Appendix~\ref{app:calibration_details}.
The final anomaly score is
\begin{equation}
S_t
=
r_{z,t}
\left[
1+\lambda_{\mathrm{gate}}\operatorname{ReLU}(d_{z,t})
\right].
\label{eq:normality_gated_score}
\end{equation}
Here, \(\lambda_{\mathrm{gate}}\) is a weighting hyperparameter controlling
the strength of discrepancy based modulation.

A positive discrepancy amplifies positive reconstruction deviations while
pushing negative deviations farther below the decision threshold, whereas a
nonpositive discrepancy leaves the calibrated reconstruction evidence
unchanged.
Thus, discrepancy from the shared normality reference acts as a conditional
modulation signal rather than an independent anomaly score.
The final prediction is
\(\widehat{y}_t=\mathbb{I}[S_t>\tau]\).

% =========================================================
% 4. Experiments
% =========================================================
\section{Experiments}
\label{sec:experiments}

We evaluate LEARN-TS on four real-world MTSAD benchmarks and analyze its
detection performance, major components, and sensitivity to key design
choices.

% =========================================================
% 4.1 Experimental Setting
% =========================================================
\subsection{Experimental Setting}
\label{sec:experimental_setting}

\paragraph{Datasets.}
We evaluate LEARN-TS on SWaT~\citep{goh2017swat},
SMD~\citep{su2019omnianomaly}, PSM~\citep{abdulaal2021psm}, and
MSL~\citep{hundman2018msl}, covering industrial control, server
monitoring, application services, and spacecraft telemetry.
SMD is evaluated entity-wise, whereas MSL uses a shared model across its
telemetry entities, with windows constructed separately within each entity.
Dataset statistics, splits, and entity-handling protocols are provided in
Appendix~\ref{app:dataset_protocol}.

\paragraph{Baselines.}
We compare LEARN-TS with 10 representative baselines:
(1) dedicated TSAD methods,
OmniAnomaly (Omni)~\citep{su2019omnianomaly} and
Anomaly Transformer (A.T.)~\citep{xu2022anomalytransformer};
(2) general time-series backbones,
TimesNet (Times)~\citep{wu2023timesnet},
PatchTST (Patch)~\citep{nie2023patchtst}, and
iTransformer (iTrans)~\citep{liu2024itransformer};
(3) language model-based methods,
GPT4TS (G4TS)~\citep{zhou2023gpt4ts},
Time-LLM (TLLM)~\citep{jin2024timellm},
LLM-Mixer (LMixer)~\citep{kowsher2025llmmixer}, and
CALF~\citep{liu2025calf}; and
(4) the pretrained general anomaly detector
DADA~\citep{shentu2025dada}.
We omit MindTS because matching its external-text setting on these
benchmarks would require additional text curation.
We use public implementations and recommended configurations when available,
with a common downstream metric and thresholding pipeline for all methods. Unless otherwise stated, LEARN-TS uses GPT-2~\citep{radford2019language} as the frozen language backbone;
backbone robustness is evaluated in Appendix~\ref{app:backbone_efficiency}.

\paragraph{Metrics.}
We report two threshold-dependent event metrics,
Range-based F1 (R-F1)~\citep{tatbul2018range} and
Affiliation F1 (Aff-F1)~\citep{huet2022affiliation}, and two
threshold-independent ranking metrics,
Area under the Precision--Recall curve
(A-PR)~\citep{saito2015precision} and
Volume under the Precision--Recall surface
(VUS-PR, denoted V-PR)~\citep{paparrizos2022vus}.
Our overlap-aware R-F1 accounts for temporal coverage and fragmented detections.
Together, these metrics assess event-level localization and anomaly score
discrimination without point adjustment~\citep{kim2022rigorous}.
Model hyperparameters and calibration parameters are fixed using
validation data. For threshold-dependent metrics, we report the R-F1-maximizing
operating point over normal-validation quantile candidates;
Aff-F1 uses the same threshold. Further details appear in
Appendices~\ref{app:dataset_protocol}
and~\ref{app:implementation_details}.
\begin{table*}[t]
\caption{
Performance comparison on four real-world multivariate time-series
anomaly detection datasets.
Results are averaged over five runs; standard deviations are
reported in Table~\ref{tab:all_methods_std}.
For each dataset--metric pair, the best and second-best results
are shown in \textbf{bold} and \underline{underlined}, respectively.
}
\label{tab:main_resultss}
\centering
\scriptsize
\setlength{\tabcolsep}{3.5pt}
\renewcommand{\arraystretch}{1.12}

\resizebox{\textwidth}{!}{%
\begin{tabular}{
    ll
    ccccccccccc
}
\toprule
Dataset
& Metric
& Omni
& A.T.
& Times
& Patch
& iTrans
& G4TS
& TLLM
& LMixer
& CALF
& DADA
& LEARN-TS \\
\midrule

\multirow{4}{*}{\textbf{SWaT}}
& A-PR
& 0.1492
& \underline{0.6829}
& 0.1341
& 0.0890
& 0.0931
& 0.0894
& 0.0846
& 0.0866
& 0.0819
& 0.5385
& \textbf{0.7578} \\

& V-PR
& 0.1371
& 0.4508
& 0.1656
& 0.0982
& 0.1008
& 0.0991
& 0.0930
& 0.0951
& 0.0907
& \underline{0.4543}
& \textbf{0.4859} \\

& R-F1
& \underline{0.1718}
& 0.1602
& 0.1593
& 0.1097
& 0.1277
& 0.1344
& 0.1190
& 0.1201
& 0.1137
& 0.0990
& \textbf{0.3102} \\

& Aff-F1
& 0.7137
& \underline{0.7413}
& \textbf{0.7940}
& 0.6376
& 0.6621
& 0.6826
& 0.6658
& 0.6737
& 0.6570
& 0.7119
& 0.7277 \\

\midrule

\multirow{4}{*}{\textbf{SMD}}
& A-PR
& 0.3713
& 0.2575
& 0.4640
& 0.4893
& 0.4726
& \textbf{0.4947}
& 0.4514
& 0.4307
& 0.4578
& 0.4754
& \underline{0.4930} \\

& V-PR
& 0.3752
& 0.3214
& 0.6184
& \underline{0.6376}
& 0.6250
& \textbf{0.6455}
& 0.5839
& 0.5662
& 0.5910
& 0.5912
& 0.5398 \\

& R-F1
& 0.1178
& 0.1930
& 0.2175
& 0.2388
& 0.2375
& 0.2427
& 0.2611
& 0.2177
& \underline{0.2725}
& 0.2683
& \textbf{0.3203} \\

& Aff-F1
& 0.6510
& 0.6818
& 0.7313
& 0.7297
& 0.7284
& 0.7153
& 0.7507
& 0.7156
& \underline{0.7553}
& 0.7543
& \textbf{0.8121} \\

\midrule

\multirow{4}{*}{\textbf{PSM}}
& A-PR
& 0.3401
& 0.3886
& 0.4013
& \underline{0.4307}
& 0.4250
& 0.4269
& 0.4169
& 0.4209
& 0.3999
& 0.4295
& \textbf{0.5700} \\

& V-PR
& 0.3622
& 0.4243
& 0.5421
& 0.5667
& \underline{0.5863}
& 0.5490
& 0.5021
& 0.5137
& 0.5136
& 0.5198
& \textbf{0.6025} \\

& R-F1
& 0.3091
& 0.2168
& 0.4451
& 0.4685
& 0.4712
& \underline{0.4896}
& 0.4665
& 0.4690
& 0.4354
& 0.4495
& \textbf{0.5256} \\

& Aff-F1
& 0.5980
& 0.4356
& \underline{0.7145}
& 0.6545
& 0.7005
& 0.7092
& 0.6873
& 0.6852
& 0.7021
& 0.6404
& \textbf{0.7765} \\

\midrule

\multirow{4}{*}{\textbf{MSL}}
& A-PR
& 0.1706
& 0.1353
& 0.2823
& 0.2696
& \underline{0.3116}
& 0.2799
& 0.2851
& 0.2826
& 0.2818
& 0.2851
& \textbf{0.3472} \\

& V-PR
& 0.2139
& 0.1742
& \underline{0.3474}
& 0.3195
& 0.3256
& 0.3253
& 0.3254
& 0.3248
& 0.3302
& 0.3259
& \textbf{0.3752} \\

& R-F1
& 0.1231
& 0.1427
& 0.2206
& 0.1584
& 0.1255
& 0.2178
& 0.2581
& \underline{0.2844}
& 0.1392
& 0.1962
& \textbf{0.2951} \\

& Aff-F1
& 0.4981
& 0.5765
& 0.6363
& 0.4137
& 0.3429
& 0.6232
& 0.6666
& \underline{0.6746}
& 0.4943
& 0.5961
& \textbf{0.7481} \\

\bottomrule
\end{tabular}%
}
\end{table*}
% =========================================================
% 4.2 Main Results (표준편차까지 있는 테이블 appendix 어디 table 에 있다 본문에 넣기)
% =========================================================
\subsection{Main Results}
\label{sec:main_results}

Table~\ref{tab:main_resultss} presents the performance comparison on four
real-world multivariate time-series anomaly detection datasets. The results
of LEARN-TS are averaged over five random seeds. It achieves the best
performance in 13 of 16 dataset--metric combinations, including R-F1 on all
four datasets and A-PR on three. The gains are most consistent on PSM and MSL,
where LEARN-TS ranks first across all four metrics. On SWaT, it leads in
A-PR, V-PR, and R-F1, while TimesNet attains the highest Aff-F1. On SMD,
LEARN-TS achieves the best R-F1 and Aff-F1, whereas GPT4TS obtains the best
A-PR and V-PR. Overall, the results indicate that LEARN-TS performs competitively across complementary ranking- and event-level metrics, although the gains vary across datasets.  

\subsection{Model Analysis}
\label{sec:model_analysis}

Table~\ref{tab:ablation_analysis} separates the effects of observation
fusion, normality guidance, and reference content using five-seed
macro averages across four datasets.
Dataset-wise results appear in Appendix~\ref{app:datasetwise_ablation}.

\paragraph{Cross-modal fusion.}
Table~\ref{tab:ablation_analysis}(a) compares removing observation
conditioning (w/o Obs.), concatenating pooled observation embeddings
with numerical representations (Concat fusion), and the proposed
gated cross-modal fusion (Full).
Removing observation conditioning reduces macro-averaged A-PR,
V-PR, and R-F1, while concatenation partially recovers performance.
Full achieves the highest A-PR, V-PR, and R-F1, whereas Concat fusion
yields slightly higher Aff-F1.
These results support the utility of the observation pathway
and selective cross-modal fusion.

\begin{table}[t]
\caption{
Ablation analysis on the four-dataset macro average.
Results are five-seed mean \(\pm\) sample standard deviation;
the best result within each panel is shown in \textbf{bold}.
}
\label{tab:ablation_analysis}
\centering
\scriptsize
\setlength{\tabcolsep}{2.7pt}
\renewcommand{\arraystretch}{1.04}

\textit{(a) Cross-modal fusion}

\vspace{0.4mm}

\begin{adjustbox}{max width=0.98\linewidth}
\begin{tabular*}{0.98\linewidth}{
    @{\extracolsep{\fill}}
    lcccc
    @{}
}
\toprule
Variant & A-PR & V-PR & R-F1 & Aff-F1 \\
\midrule

w/o Obs.
& 0.5119 \(\pm\) 0.0153
& 0.4746 \(\pm\) 0.0041
& 0.3333 \(\pm\) 0.0064
& 0.7636 \(\pm\) 0.0045 \\

Concat fusion
& 0.5287 \(\pm\) 0.0284
& 0.4867 \(\pm\) 0.0151
& 0.3500 \(\pm\) 0.0054
& \textbf{0.7688 \(\pm\) 0.0037} \\

Full
& \textbf{0.5420 \(\pm\) 0.0060}
& \textbf{0.5008 \(\pm\) 0.0105}
& \textbf{0.3628 \(\pm\) 0.0119}
& 0.7661 \(\pm\) 0.0066 \\

\bottomrule
\end{tabular*}
\end{adjustbox}

\vspace{0.6mm}

\textit{(b) Normality alignment and scoring}

\vspace{0.4mm}

\begin{adjustbox}{max width=0.98\linewidth}
\begin{tabular*}{0.98\linewidth}{
    @{\extracolsep{\fill}}
    lcccccc
    @{}
}
\toprule
Variant & Align. & Score & A-PR & V-PR & R-F1 & Aff-F1 \\
\midrule

No normality
& \xmark
& \xmark
& 0.3865 \(\pm\) 0.0062
& 0.4108 \(\pm\) 0.0036
& \textbf{0.3654 \(\pm\) 0.0109}
& \textbf{0.7711 \(\pm\) 0.0022} \\

Scoring only
& \xmark
& \cmark
& 0.3801 \(\pm\) 0.0110
& 0.4136 \(\pm\) 0.0078
& 0.3531 \(\pm\) 0.0133
& 0.7644 \(\pm\) 0.0058 \\

Alignment only
& \cmark
& \xmark
& 0.3831 \(\pm\) 0.0047
& 0.4106 \(\pm\) 0.0023
& 0.3579 \(\pm\) 0.0089
& 0.7705 \(\pm\) 0.0023 \\

Full
& \cmark
& \cmark
& \textbf{0.5420 \(\pm\) 0.0060}
& \textbf{0.5008 \(\pm\) 0.0105}
& 0.3628 \(\pm\) 0.0119
& 0.7661 \(\pm\) 0.0066 \\

\bottomrule
\end{tabular*}
\end{adjustbox}

\vspace{0.6mm}

\textit{(c) Normality-reference control}

\vspace{0.4mm}

\begin{adjustbox}{max width=0.98\linewidth}
\begin{tabular*}{0.98\linewidth}{
    @{\extracolsep{\fill}}
    lcccc
    @{}
}
\toprule
Normality reference & A-PR & V-PR & R-F1 & Aff-F1 \\
\midrule

Random semantic-free
& 0.5360 \(\pm\) 0.0113
& 0.4965 \(\pm\) 0.0104
& 0.3596 \(\pm\) 0.0052
& 0.7623 \(\pm\) 0.0045 \\

Dataset-agnostic semantic
& \textbf{0.5420 \(\pm\) 0.0060}
& \textbf{0.5008 \(\pm\) 0.0105}
& \textbf{0.3628 \(\pm\) 0.0119}
& \textbf{0.7661 \(\pm\) 0.0066} \\

\bottomrule
\end{tabular*}
\end{adjustbox}
\end{table}

% \paragraph{Normality alignment and scoring.}
% Table~\ref{tab:ablation_analysis}(b) shows that jointly applying normality
% alignment and discrepancy-guided scoring substantially improves A-PR and
% V-PR, whereas R-F1 and Aff-F1 are less affected.
% Neither component alone reproduces the ranking gains of their joint use,
% supporting their complementary roles.
% Overall, the shared normality reference primarily improves
% anomaly-score discrimination rather than uniformly improving
% event-level detection.
\paragraph{Normality alignment and scoring.}
Under the semantic-reference configuration,
Table~\ref{tab:ablation_analysis}(b) shows that jointly applying
normality alignment and discrepancy-guided scoring improves
macro-averaged A-PR and V-PR over either component alone or
no normality guidance.
Neither component alone reproduces the ranking gains of their
joint use, supporting their complementary roles.
Mean R-F1 and Aff-F1 remain slightly below the no-normality
variant, indicating that the joint mechanism primarily benefits
score ranking rather than uniformly improving event-level detection.

\paragraph{Normality-reference control.}
With the alignment-and-scoring pipeline held fixed,
Table~\ref{tab:ablation_analysis}(c) compares the dataset-agnostic normality reference with a random semantic-free reference. The semantic reference achieves slightly
higher means across all four metrics, suggesting a modest
additional benefit from reference content.
Reference construction details are provided in
Appendix~\ref{app:normality_references}.

\begin{figure*}[t]
    \centering
    \includegraphics[width=0.95\textwidth]
    {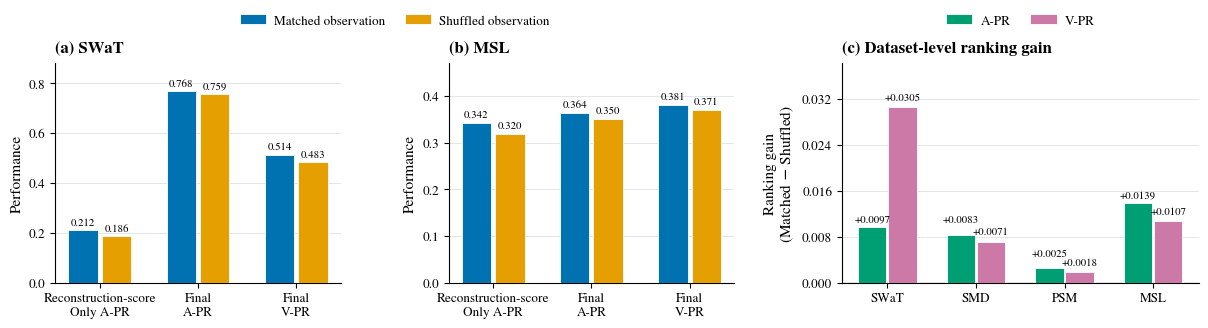}
    \caption{
Effect of observation--window correspondence:
(a)--(b) matched vs.\ shuffled observations on SWaT and MSL;
(c) A-PR and V-PR gains across all four datasets.
    }
    \label{fig:matched_shuffled}
\end{figure*}

\paragraph{Observation--window correspondence.}
Figure~\ref{fig:matched_shuffled} compares matched observations with
representations shuffled across windows while preserving the architecture.
Using seed 42, matched observations improve reconstruction-only A-PR
and final A-PR and V-PR across all four datasets, supporting the importance
of correspondence between observation semantics and the underlying window.

% \paragraph{Observation--window correspondence.}
% Figure~\ref{fig:matched_shuffled} compares matched observations with
% representations shuffled across windows, preserving the architecture.
% Using seed 42, matched observations improve reconstruction-only A-PR
% and final A-PR and V-PR across all four datasets.
% A complementary LLM replacement control using the same seed replaces
% both language-derived representations with MLP outputs.
% GPT-2 Full achieves higher A-PR on three datasets and V-PR on all four
% than MLP (w/o LLM), while threshold-dependent metrics are mixed
% (Appendix~\ref{app:obs_encoder_control}).

% 화질이 안 좋음

\begin{figure}[t]
    \centering
    \includegraphics[width=1\linewidth]
    {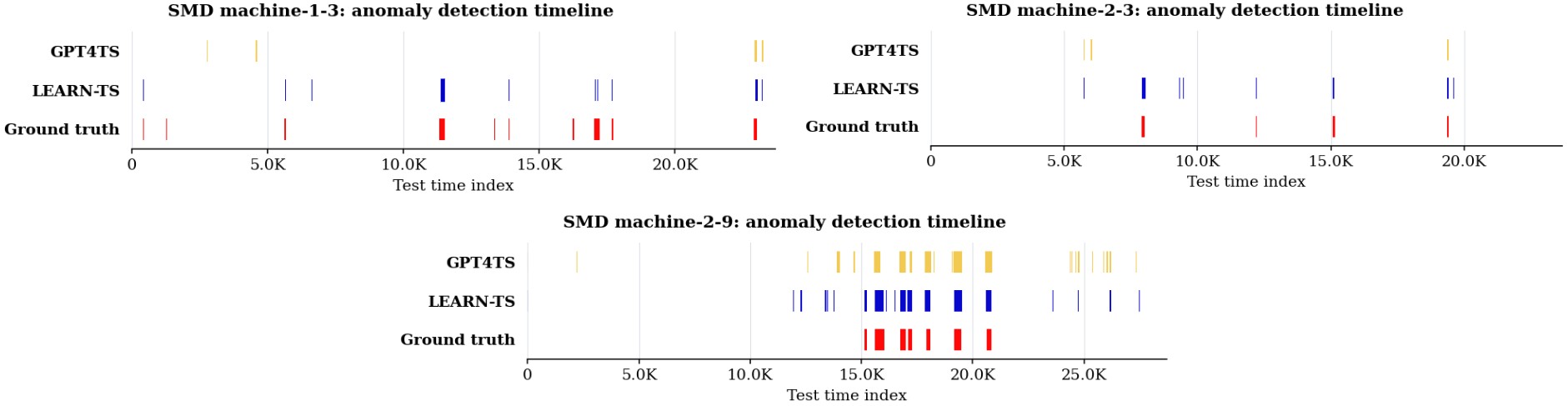}
    \caption{
    Event-level predictions on three SMD entities:
    Machine 1-3 with intermediate event density,
    Machine 2-3 with sparse events, and
    Machine 2-9 with dense events.
    % LEARN-TS generally achieves closer temporal overlap with the
    % ground truth events than GPT4TS.
    }
    \label{fig:smd_event_timelines}
\end{figure}

\paragraph{Qualitative analysis.}
We compare with GPT4TS, the top SMD baseline in A-PR and V-PR.
Figure~\ref{fig:smd_event_timelines} compares event-level predictions on
three SMD entities with different anomaly densities.
Here, LEARN-TS predictions often overlap more closely with
labeled events, while GPT4TS misses some events in the sparse case and
produces detections outside the main event cluster in the dense case.

\paragraph{Sensitivity analysis.}

\begin{wrapfigure}[17]{r}{0.52\columnwidth}
    \vspace{-1.2\baselineskip}
    \centering
    \includegraphics[width=\linewidth]
    {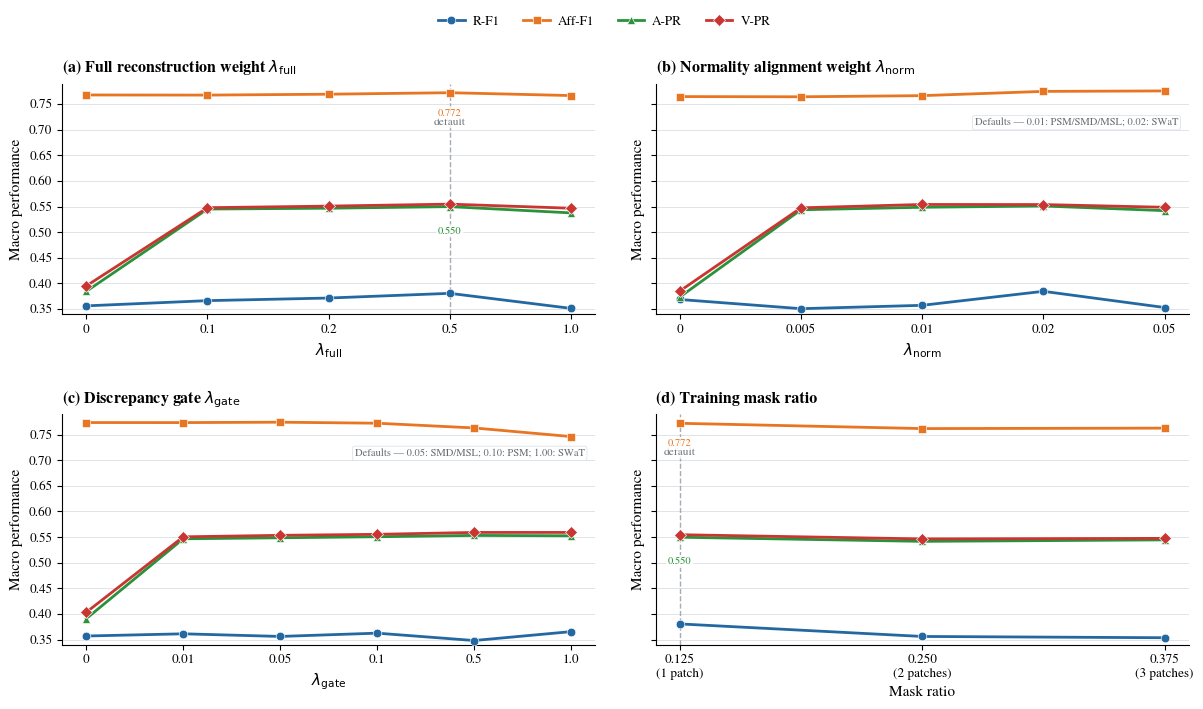}
    \caption{
        Sensitivity to
        (a) $\lambda_{\mathrm{full}}$,
        (b) $\lambda_{\mathrm{norm}}$,
        (c) $\lambda_{\mathrm{gate}}$, and
        (d) the number of masked training patches.
        Results are four-dataset macro averages using a fixed seed.
    }
    \label{fig:sensitivity}
    \vspace{-0.5\baselineskip}
\end{wrapfigure}

Figure~\ref{fig:sensitivity} summarizes fixed-seed sensitivity trends.
Removing the full-window objective or normality guidance reduces macro
ranking performance, whereas performance is generally more stable across
positive weight values.
A nonzero discrepancy gate improves A-PR and V-PR over reconstruction-only
scoring, although its effects vary across datasets and metrics.
Single-patch masking yields the most consistent aggregate performance among
the evaluated masking configurations.
Overall, the selected values provide favorable aggregate trade-offs across
the four reported metrics.
Dataset-wise results and validation-selected settings are provided in
Appendices~\ref{app:datasetwise_sensitivity}
and~\ref{app:implementation_details}, respectively.

% Machine 1-3 — intermediate-event 사례
% Machine 2-9 — dense-event 사례
% Machine 2-3 — sparse-event 사례

% =========================================================
% 5. Conclusion
% =========================================================
\section{Conclusion}
\label{sec:conclusion}
% We proposed LEARN-TS, a semantically guided MTSAD framework that combines
% observation-conditioned patch masked reconstruction with a window-independent
% normality reference without requiring time associated paired external text.
We studied MTSAD without requiring time-associated paired external text.
We proposed LEARN-TS, a semantically guided framework that combines
observation-conditioned patch-masked reconstruction with a
window-independent normality reference. Across four real-world benchmarks, LEARN-TS ranks first in 13 of 16 primary dataset--metric comparisons.
Controlled analyses support the importance of observation--window
correspondence and show that normality alignment and discrepancy-guided
scoring benefit anomaly-score ranking, with modest gains from the
dataset-agnostic semantic reference over a random semantic-free alternative.
Future work will investigate expert-validated sensor-level normality
constraints and reduce the cost of patch-wise inference.

\section*{AI Use Statement}
We used AI assistants to support discussions of experimental
evaluation and results, assist with literature discovery, and
improve manuscript text and LaTeX tables.
The authors reviewed all AI-assisted material, made the final
research decisions, and take full responsibility for the content.

% =========================================================
% References
% =========================================================
\bibliography{refs}
\bibliographystyle{iclr2027_conference}

% =========================================================
% Appendix
% =========================================================
\clearpage
\appendix

% =========================================================
% Appendix A. Dataset Statistics and Evaluation Protocol
% =========================================================
\section{Dataset Statistics and Evaluation Protocol}
\label{app:dataset_protocol}

Table~\ref{tab:dataset_statistics} summarizes the datasets used in our
experiments.
The reported train and test sizes correspond to the official splits before
our internal training--validation division.
For SMD and MSL, sequence lengths are summed over all entities, and the
anomaly ratio denotes the percentage of anomalous timestamps in the official
test split.

\paragraph{Preprocessing and validation split.}
We preserve the official test splits without modification.
Each official training sequence is chronologically divided into the first
80\% for training and the remaining 20\% for validation before window
construction.
For SMD and MSL, this split is performed independently within each entity.
Feature-wise standardization statistics are estimated exclusively from the
resulting training portions and applied unchanged to validation and test data.
Windows have length \(L=128\), and no window crosses an entity boundary.
Normal validation data are used for score calibration, hyperparameter
selection, and quantile-based threshold candidate construction.
For threshold-dependent metrics, we report the operating point that
maximizes R-F1 over these candidates and evaluate Aff-F1 at the same
threshold.

\paragraph{Entity-wise protocol.}
For SMD, a separate model is trained and evaluated for each of the
28 machines, and machine-level metrics are macro-averaged.
For MSL, a shared model is trained on windows constructed within
each of the 27 telemetry entities. Inference is performed separately
for each entity, after which scores and labels are concatenated.
Concatenated MSL processing is also used in the public
OmniAnomaly implementation~\citep{su2019omnianomaly}.
Smoothing and metric computation are applied to the pooled sequence
without explicit boundary handling.
SWaT and PSM are each treated as a single multivariate sequence.

\paragraph{Timeline aggregation and evaluation.}
Window-level outputs are mapped back to the original timeline.
Reconstruction errors are retained only from inference passes in which the
corresponding temporal patch is masked.
Window-level normality discrepancies are broadcast to their covered
timestamps, and overlapping contributions are aggregated using mean
aggregation for SWaT and MSL and maximum aggregation for SMD and PSM.
All methods are evaluated using the same score-processing, metric, and
thresholding pipeline.

\paragraph{Baseline evaluation.}
For all baselines, we preserve each method's original anomaly-scoring
procedure and evaluate the resulting anomaly-score sequences using the same
downstream score-processing, metric, and threshold selection
pipeline adopted for LEARN-TS.
For DADA, we use the officially released pretrained checkpoint in its
zero-shot setting without target domain fine tuning, while preserving its
official preprocessing and inference procedures.

\paragraph{Metric implementation.}
\label{app:metrics}

Before evaluation, score smoothing is applied uniformly to all methods
using a window of 30 timestamps for SWaT and 10 for the other datasets.
R-F1 uses an overlap-aware range configuration that accounts for temporal
coverage and fragmented detections, while Aff-F1 evaluates affiliation
between predicted and ground-truth events.
% Both metrics are computed from binary predictions at the operating point
% described above.

A-PR is computed from continuous anomaly scores and the original point-wise
labels.
VUS-PR is computed using the reference implementation of
\citet{paparrizos2022vus}
(\texttt{RangeAUC\_volume\_opt}), with 250 score thresholds and temporal
buffer lengths up to 200 timestamps.
A-PR and VUS-PR are threshold-independent and do not use the selected
operating point.
No point adjustment is applied to any metric.
Additional evaluator settings are provided in the anonymized code.

\section{Implementation Details}
\label{app:implementation_details}
Exact implementation constants and complete configuration files are provided
in the anonymized code release.

\begin{table*}[t]
\caption{
Dataset statistics before the internal training--validation split.
For SMD and MSL, sequence lengths are summed over all entities.
}
\label{tab:dataset_statistics}
\centering
\small
\setlength{\tabcolsep}{7pt}
\renewcommand{\arraystretch}{1.08}
\begin{tabular}{llrrrrr}
\toprule
Dataset & Domain & Entities & Variables
& Official train & Official test & Anomaly (\%) \\
\midrule
SWaT & Industrial control & 1  & 51 & 496,800 & 449,919 & 11.98 \\
SMD  & Server monitoring  & 28 & 38 & 708,405 & 708,420 & 4.16 \\
PSM  & Application server & 1  & 25 & 132,481 & 87,841  & 27.75 \\
MSL  & Spacecraft telemetry & 27 & 55 & 58,317 & 73,729 & 10.72 \\
\bottomrule
\end{tabular}
\end{table*}

\paragraph{Data preprocessing.}
Feature-wise standardization statistics are computed exclusively from the
normal training split and subsequently applied without modification to the
validation and test splits. Each multivariate time series is segmented into
input windows of length \(L=128\). Each window is divided into
\(P=8\) non-overlapping temporal patches of length \(\ell=16\).

\begin{table}[t]
\caption{Common model and optimization settings used across all datasets.}
\label{tab:common_hyperparameters}
\centering
\small
\setlength{\tabcolsep}{6pt}
\renewcommand{\arraystretch}{1.1}
\begin{tabular}{lc}
\toprule
Hyperparameter & Value \\
\midrule
Input window length \(L\)          & 128 \\
Patch length \(\ell\)              & 16 \\
Number of patches \(P\)            & 8 \\
Hidden dimension \(d\)             & 768 \\
Time-series encoder layers         & 3 \\
Cross-modal fusion layers          & 2 \\
Attention heads                    & 8 \\
Batch size                         & 32 \\
Optimizer                          & AdamW \\
Weight decay                       & \(1\times10^{-4}\) \\
Gradient clipping                  & 1.0 \\
Initial learning rate & \(5\times10^{-5}\)\({}^{\dagger}\) \\
Minimum learning rate              & \(1\times10^{-6}\) \\
Early-stopping patience            & 5 epochs \\
\(\lambda_{\mathrm{full}}\)        & 0.5 \\
\bottomrule
\end{tabular}

\vspace{2pt}
\makebox[\linewidth][c]{%
  \footnotesize
  \({}^{\dagger}\) PSM uses an initial learning rate of \(2\times10^{-5}\).%
}
\end{table}

\begin{table}[t]
\caption{Dataset-specific loss and scoring configurations.}
\label{tab:dataset_hyperparameters}
\centering
\small
\setlength{\tabcolsep}{7pt}
\renewcommand{\arraystretch}{1.1}
\begin{tabular}{lcc}
\toprule
Dataset & \(\lambda_{\mathrm{norm}}\) & \(\lambda_{\mathrm{gate}}\) \\
\midrule
SWaT & 0.02 & 1.00 \\
SMD  & 0.01 & 0.05 \\
PSM  & 0.01 & 0.1 \\
MSL  & 0.01 & 0.05 \\
\bottomrule
\end{tabular}
\end{table}

\begin{table}[t]
\caption{
Dataset-specific channel error and overlapping-window aggregation rules.
}
\label{tab:score_aggregation}
\centering
\small
\setlength{\tabcolsep}{5pt}
\renewcommand{\arraystretch}{1.1}
\begin{tabular}{lcc}
\toprule
Dataset
& Channel aggregation \(\mathcal G_{\mathcal D}\)
& Window overlap \\
\midrule
SWaT & Mean over all variables  & Mean \\
SMD  & Top-\(k\) mean           & Maximum \\
PSM  & Top-\(k\) mean           & Maximum \\
MSL  & Target telemetry channel & Mean \\
\bottomrule
\end{tabular}
\end{table}

\paragraph{Patch masking.}
During training and validation, one temporal patch is sampled uniformly and
masked at the same temporal location across all variables. At inference,
stochastic masking is replaced by deterministic exhaustive patch-wise
masking. Specifically, each of the \(P\) patches is masked once, resulting in
eight single patch masked forward passes for each input window.
Reconstruction errors are retained only for timestamps belonging to the
currently masked patch. The retained errors from all forward passes are then
combined to construct dense reconstruction evidence over the complete input
window.

\paragraph{Model configuration.}
The hidden dimension is set to \(d=768\). The time-series encoder consists of
three layers, and the gated cross-modal fusion module consists of two layers.
Each encoder and fusion layer uses eight attention heads. The text encoder is
kept frozen, whereas the time-series encoder, text projection module,
cross-modal fusion module, and reconstruction head are optimized during
training.
The common model and optimization settings used across all datasets are
summarized in Table~\ref{tab:common_hyperparameters}.

\paragraph{Optimization.}
All trainable components are optimized using AdamW with a batch size of 32,
weight decay of \(10^{-4}\), and gradient clipping with a maximum norm of
1.0. 
% The initial learning rate is \(5\times10^{-5}\) for all datasets.
The initial learning rate is \(5\times10^{-5}\) for SWaT, SMD, and MSL,
and \(2\times10^{-5}\) for PSM.
We apply cosine annealing with
a minimum learning rate of \(10^{-6}\). Training is terminated when the
validation loss does not improve for five consecutive epochs. All experiments were conducted on an NVIDIA L40S GPU
with 48 GB of GPU memory.

\paragraph{Loss configuration.}
We use \(\lambda_{\mathrm{full}}=0.5\) for all datasets. The
normality-alignment weight is set to \(\lambda_{\mathrm{norm}}=0.02\) for
SWaT and \(\lambda_{\mathrm{norm}}=0.01\) for SMD, PSM, and MSL.
The dataset-specific configurations are summarized in
Table~\ref{tab:dataset_hyperparameters}.

\paragraph{Dataset-specific score aggregation.}
The operator \(\mathcal G_{\mathcal D}\) converts the channel-wise
reconstruction errors in Equation~\ref{eq:reconstruction_evidence} into a
scalar score using the dataset-specific rule in
Table~\ref{tab:score_aggregation}.
For top-\(k\) aggregation, the \(k\) largest channel errors are averaged.
Scores contributed by overlapping windows are aggregated using either their
mean or maximum.
The same window-overlap rule is used to aggregate the broadcast normality
discrepancies.

\paragraph{Score calibration and anomaly prediction.}
\label{app:calibration_details}

Reconstruction evidence and normality discrepancy are calibrated
independently using their timestamp-level distributions on normal validation
data.
For \(a\in\{r,d\}\), let \(\mathcal Q_N^{(a)}\) denote the corresponding
normal validation distribution.
The calibrated score is
\begin{equation}
a_{z,t}
=
\operatorname{clip}\!\left(
\frac{
a_t-\operatorname{median}(\mathcal Q_N^{(a)})
}{
s_N^{(a)}
},
-\kappa,\kappa
\right),
\qquad
a\in\{r,d\},
\label{eq:robust_calibration_details}
\end{equation}
where \(s_N^{(a)}\) is a positive robust scale computed from
\(\mathcal Q_N^{(a)}\).
In implementation, the scale takes the maximum of normalized
median absolute deviation, central quantile range, standard deviation, and
positive-floor estimates.
This composite definition prevents near-degenerate validation score
distributions from producing excessively large calibrated values.

The calibrated scores are combined according to
Equation~\ref{eq:normality_gated_score}.
The discrepancy-gating coefficient is selected using validation data.
Quantile-based threshold candidates are constructed from normal validation
scores, with the final operating point selected as described in
Appendix~\ref{app:metrics}.

\paragraph{Random seeds and reporting.}
The main results are averaged over five independent runs with
different random seeds. For each trainable method, the same
five-seed evaluation protocol is used.
The module ablations in Table~\ref{tab:ablation_analysis} use five
seeds, whereas sensitivity analyses and the observation
correspondence control use a single fixed seed (42).

Dataset-level results are reported as the mean and sample standard deviation
over five random seeds. For trainable methods, the seeds control model
initialization and stochastic training. DADA uses its fixed officially
released pretrained checkpoint and is independently evaluated under five
random seeds, which control the stochastic components of its inference
pipeline. For four-dataset macro results, the four datasets are first
averaged with equal weight within each seed, after which the final mean and
sample standard deviation are computed over the five seed-level averages.
Table~\ref{tab:all_methods_std} reports five-run summary statistics
for each method, dataset, and primary metric.
Per-seed raw results are provided in the supplementary result files.
Evaluation labels are not used for model selection or model
hyperparameter tuning. For metric reporting, they select the
R-F1-maximizing threshold from validation-derived candidates;
Aff-F1 uses the same threshold.

\begin{table*}[t]
\caption{
Performance of all methods on the four datasets over five independent runs.
Results are reported as mean \(\pm\) sample standard deviation.
For each dataset--metric pair, the best and second-best mean results are
shown in \textbf{bold} and \underline{underlined}, respectively.
}
\label{tab:all_methods_std}
\centering
\scriptsize
\setlength{\tabcolsep}{8pt}
\renewcommand{\arraystretch}{0.98}

\begin{tabular*}{\textwidth}{
    @{\extracolsep{\fill}}
    lcccc
    @{}
}
\toprule
Method & A-PR & V-PR & R-F1 & Aff-F1 \\
\midrule

\multicolumn{5}{l}{\textit{(a) SWaT}} \\
\addlinespace[2pt]

OmniAnomaly
& 0.1492 \(\pm\) 0.0559
& 0.1371 \(\pm\) 0.0321
& \underline{0.1718 \(\pm\) 0.1140}
& 0.7137 \(\pm\) 0.0041 \\

Anomaly Transformer
& \underline{0.6829 \(\pm\) 0.0193}
& 0.4508 \(\pm\) 0.0077
& 0.1602 \(\pm\) 0.0187
& \underline{0.7413 \(\pm\) 0.0149} \\

TimesNet
& 0.1341 \(\pm\) 0.0018
& 0.1656 \(\pm\) 0.0021
& 0.1593 \(\pm\) 0.0057
& \textbf{0.7940 \(\pm\) 0.0091} \\

PatchTST
& 0.0890 \(\pm\) 0.0002
& 0.0982 \(\pm\) 0.0002
& 0.1097 \(\pm\) 0.0062
& 0.6376 \(\pm\) 0.0026 \\

iTransformer
& 0.0931 \(\pm\) 0.0010
& 0.1008 \(\pm\) 0.0005
& 0.1277 \(\pm\) 0.0029
& 0.6621 \(\pm\) 0.0025 \\

GPT4TS
& 0.0894 \(\pm\) 0.0002
& 0.0991 \(\pm\) 0.0003
& 0.1344 \(\pm\) 0.0015
& 0.6826 \(\pm\) 0.0056 \\

Time-LLM
& 0.0846 \(\pm\) 0.0003
& 0.0930 \(\pm\) 0.0003
& 0.1190 \(\pm\) 0.0021
& 0.6658 \(\pm\) 0.0016 \\

LLM-Mixer
& 0.0866 \(\pm\) 0.0004
& 0.0951 \(\pm\) 0.0004
& 0.1201 \(\pm\) 0.0033
& 0.6737 \(\pm\) 0.0009 \\

CALF
& 0.0819 \(\pm\) 0.0003
& 0.0907 \(\pm\) 0.0004
& 0.1137 \(\pm\) 0.0016
& 0.6570 \(\pm\) 0.0062 \\

DADA
& 0.5385 \(\pm\) 0.0042
& \underline{0.4543 \(\pm\) 0.0031}
& 0.0990 \(\pm\) 0.0061
& 0.7119 \(\pm\) 0.0076 \\

LEARN-TS
& \textbf{0.7578 \(\pm\) 0.0117}
& \textbf{0.4859 \(\pm\) 0.0346}
& \textbf{0.3102 \(\pm\) 0.0436}
& 0.7277 \(\pm\) 0.0069 \\

\midrule
\multicolumn{5}{l}{\textit{(b) SMD}} \\
\addlinespace[2pt]

OmniAnomaly
& 0.3713 \(\pm\) 0.0123
& 0.3752 \(\pm\) 0.0083
& 0.1178 \(\pm\) 0.0060
& 0.6510 \(\pm\) 0.0087 \\

Anomaly Transformer
& 0.2575 \(\pm\) 0.0110
& 0.3214 \(\pm\) 0.0050
& 0.1930 \(\pm\) 0.0060
& 0.6818 \(\pm\) 0.0183 \\

TimesNet
& 0.4640 \(\pm\) 0.0048
& 0.6184 \(\pm\) 0.0027
& 0.2175 \(\pm\) 0.0049
& 0.7313 \(\pm\) 0.0052 \\

PatchTST
& 0.4893 \(\pm\) 0.0009
& \underline{0.6376 \(\pm\) 0.0009}
& 0.2388 \(\pm\) 0.0022
& 0.7297 \(\pm\) 0.0017 \\

iTransformer
& 0.4726 \(\pm\) 0.0008
& 0.6250 \(\pm\) 0.0010
& 0.2375 \(\pm\) 0.0028
& 0.7284 \(\pm\) 0.0014 \\

GPT4TS
& \textbf{0.4947 \(\pm\) 0.0006}
& \textbf{0.6455 \(\pm\) 0.0008}
& 0.2427 \(\pm\) 0.0066
& 0.7153 \(\pm\) 0.0039 \\

Time-LLM
& 0.4514 \(\pm\) 0.0008
& 0.5839 \(\pm\) 0.0013
& 0.2611 \(\pm\) 0.0026
& 0.7507 \(\pm\) 0.0020 \\

LLM-Mixer
& 0.4307 \(\pm\) 0.0486
& 0.5662 \(\pm\) 0.0445
& 0.2177 \(\pm\) 0.0216
& 0.7156 \(\pm\) 0.0231 \\

CALF
& 0.4578 \(\pm\) 0.0013
& 0.5910 \(\pm\) 0.0009
& \underline{0.2725 \(\pm\) 0.0010}
& \underline{0.7553 \(\pm\) 0.0005} \\

DADA
& 0.4754 \(\pm\) 0.0030
& 0.5912 \(\pm\) 0.0034
& 0.2683 \(\pm\) 0.0015
& 0.7543 \(\pm\) 0.0016 \\

LEARN-TS
& \underline{0.4930 \(\pm\) 0.0056}
& 0.5398 \(\pm\) 0.0037
& \textbf{0.3203 \(\pm\) 0.0033}
& \textbf{0.8121 \(\pm\) 0.0056} \\

\midrule
\multicolumn{5}{l}{\textit{(c) PSM}} \\
\addlinespace[2pt]

OmniAnomaly
& 0.3401 \(\pm\) 0.0392
& 0.3622 \(\pm\) 0.0224
& 0.3091 \(\pm\) 0.0976
& 0.5980 \(\pm\) 0.0703 \\

Anomaly Transformer
& 0.3886 \(\pm\) 0.0527
& 0.4243 \(\pm\) 0.0449
& 0.2168 \(\pm\) 0.0308
& 0.4356 \(\pm\) 0.0466 \\

TimesNet
& 0.4013 \(\pm\) 0.0046
& 0.5421 \(\pm\) 0.0089
& 0.4451 \(\pm\) 0.0097
& \underline{0.7145 \(\pm\) 0.0046} \\

PatchTST
& \underline{0.4307 \(\pm\) 0.0027}
& 0.5667 \(\pm\) 0.0031
& 0.4685 \(\pm\) 0.0045
& 0.6545 \(\pm\) 0.0162 \\

iTransformer
& 0.4250 \(\pm\) 0.0024
& \underline{0.5863 \(\pm\) 0.0027}
& 0.4712 \(\pm\) 0.0075
& 0.7005 \(\pm\) 0.0047 \\

GPT4TS
& 0.4269 \(\pm\) 0.0021
& 0.5490 \(\pm\) 0.0014
& \underline{0.4896 \(\pm\) 0.0072}
& 0.7092 \(\pm\) 0.0133 \\

Time-LLM
& 0.4169 \(\pm\) 0.0011
& 0.5021 \(\pm\) 0.0032
& 0.4665 \(\pm\) 0.0023
& 0.6873 \(\pm\) 0.0037 \\

LLM-Mixer
& 0.4209 \(\pm\) 0.0007
& 0.5137 \(\pm\) 0.0022
& 0.4690 \(\pm\) 0.0029
& 0.6852 \(\pm\) 0.0036 \\

CALF
& 0.3999 \(\pm\) 0.0005
& 0.5136 \(\pm\) 0.0031
& 0.4354 \(\pm\) 0.0031
& 0.7021 \(\pm\) 0.0001 \\

DADA
& 0.4295 \(\pm\) 0.0025
& 0.5198 \(\pm\) 0.0024
& 0.4495 \(\pm\) 0.0089
& 0.6404 \(\pm\) 0.0103 \\

LEARN-TS
& \textbf{0.5700 \(\pm\) 0.0100}
& \textbf{0.6025 \(\pm\) 0.0094}
& \textbf{0.5256 \(\pm\) 0.0064}
& \textbf{0.7765 \(\pm\) 0.0103} \\

\midrule
\multicolumn{5}{l}{\textit{(d) MSL}} \\
\addlinespace[2pt]

OmniAnomaly
& 0.1706 \(\pm\) 0.0170
& 0.2139 \(\pm\) 0.0253
& 0.1231 \(\pm\) 0.0189
& 0.4981 \(\pm\) 0.0435 \\

Anomaly Transformer
& 0.1353 \(\pm\) 0.0062
& 0.1742 \(\pm\) 0.0056
& 0.1427 \(\pm\) 0.0084
& 0.5765 \(\pm\) 0.0201 \\

TimesNet
& 0.2823 \(\pm\) 0.0044
& \underline{0.3474 \(\pm\) 0.0046}
& 0.2206 \(\pm\) 0.0081
& 0.6363 \(\pm\) 0.0360 \\

PatchTST
& 0.2696 \(\pm\) 0.0090
& 0.3195 \(\pm\) 0.0095
& 0.1584 \(\pm\) 0.0257
& 0.4137 \(\pm\) 0.0568 \\

iTransformer
& \underline{0.3116 \(\pm\) 0.0048}
& 0.3256 \(\pm\) 0.0035
& 0.1255 \(\pm\) 0.0088
& 0.3429 \(\pm\) 0.0185 \\

GPT4TS
& 0.2799 \(\pm\) 0.0024
& 0.3253 \(\pm\) 0.0024
& 0.2178 \(\pm\) 0.0237
& 0.6232 \(\pm\) 0.0361 \\

Time-LLM
& 0.2851 \(\pm\) 0.0003
& 0.3254 \(\pm\) 0.0002
& 0.2581 \(\pm\) 0.0028
& 0.6666 \(\pm\) 0.0010 \\

LLM-Mixer
& 0.2826 \(\pm\) 0.0057
& 0.3248 \(\pm\) 0.0029
& \underline{0.2844 \(\pm\) 0.0199}
& \underline{0.6746 \(\pm\) 0.0157} \\

CALF
& 0.2818 \(\pm\) 0.0004
& 0.3302 \(\pm\) 0.0004
& 0.1392 \(\pm\) 0.0009
& 0.4943 \(\pm\) 0.0031 \\

DADA
& 0.2851 \(\pm\) 0.0084
& 0.3259 \(\pm\) 0.0085
& 0.1962 \(\pm\) 0.0164
& 0.5961 \(\pm\) 0.0331 \\

LEARN-TS
& \textbf{0.3472 \(\pm\) 0.0184}
& \textbf{0.3752 \(\pm\) 0.0053}
& \textbf{0.2951 \(\pm\) 0.0023}
& \textbf{0.7481 \(\pm\) 0.0282} \\

\bottomrule
\end{tabular*}
\end{table*}

\section{Dataset-Wise Ablation Results}
\label{app:datasetwise_ablation}

The main paper reports five-seed macro averaged ablation results for a
compact comparison of the major components.
Here, we provide dataset-wise results for cross-modal fusion, normality
alignment and scoring, and semantic-reference construction to examine
whether their effects are consistent across benchmarks.

% ---------------------------------------------------------
% Cross-modal fusion
% ---------------------------------------------------------
\subsection{Cross-Modal Fusion}
\label{app:datasetwise_cross_modal_fusion}

Table~\ref{tab:datasetwise_cma} reports the dataset-wise cross-modal fusion
ablation. We compare three architectural configurations:
\textit{w/o Obs.}, which removes observation-text conditioning;
\textit{Concat fusion}, which replaces gated cross-attention with pooled
text concatenation; and
\textit{LEARN-TS (Full)}, which uses gated cross-modal fusion with the
corresponding observation representation.

The full model improves A-PR, V-PR, and R-F1 substantially on SWaT and yields
the highest ranking performance on SMD. On PSM, the full model provides a
clear improvement in R-F1, although the simpler fusion variants obtain higher
ranking metrics. The differences are smaller on MSL, where the three
configurations perform comparably. These results indicate that the benefit of
gated cross-modal fusion is dataset dependent, with its strongest gains
appearing on benchmarks where observation semantics provide complementary
information for reconstruction.
The observation--window correspondence control is analyzed separately in
Figure~\ref{fig:matched_shuffled}.

\begin{table*}[t]
\caption{
Dataset-wise cross-modal fusion ablation.
Results are reported as mean \(\pm\) sample standard deviation over five
random seeds. The best result in each dataset--metric column is shown in
\textbf{bold}.
}
\label{tab:datasetwise_cma}
\centering
\footnotesize
\setlength{\tabcolsep}{6pt}
\renewcommand{\arraystretch}{1.05}
\begin{tabular}{llcccc}
\toprule
Dataset & Variant & A-PR & V-PR & R-F1 & Aff-F1 \\
\midrule
SWaT
& w/o Obs.
& 0.6552 $\pm$ 0.0652
& 0.3727 $\pm$ 0.0193
& 0.2446 $\pm$ 0.0110
& 0.7047 $\pm$ 0.0012 \\
& Concat fusion
& 0.7001 $\pm$ 0.1014
& 0.4170 $\pm$ 0.0508
& 0.2911 $\pm$ 0.0189
& 0.7270 $\pm$ 0.0090 \\
& LEARN-TS (Full)
& \textbf{0.7578 $\pm$ 0.0117}
& \textbf{0.4859 $\pm$ 0.0346}
& \textbf{0.3102 $\pm$ 0.0436}
& \textbf{0.7277 $\pm$ 0.0069} \\
\midrule

SMD
& w/o Obs.
& 0.4650 $\pm$ 0.0074
& 0.5266 $\pm$ 0.0047
& 0.3222 $\pm$ 0.0070
& 0.8137 $\pm$ 0.0102 \\
& Concat fusion
& 0.4829 $\pm$ 0.0122
& 0.5380 $\pm$ 0.0098
& \textbf{0.3262 $\pm$ 0.0057}
& \textbf{0.8174 $\pm$ 0.0084} \\
& LEARN-TS (Full)
& \textbf{0.4930 $\pm$ 0.0056}
& \textbf{0.5398 $\pm$ 0.0037}
& 0.3203 $\pm$ 0.0033
& 0.8121 $\pm$ 0.0056 \\
\midrule

PSM
& w/o Obs.
& 0.5767 $\pm$ 0.0047
& \textbf{0.6203 $\pm$ 0.0049}
& 0.4613 $\pm$ 0.0138
& \textbf{0.7783 $\pm$ 0.0110} \\
& Concat fusion
& \textbf{0.5809 $\pm$ 0.0063}
& 0.6186 $\pm$ 0.0035
& 0.4768 $\pm$ 0.0102
& 0.7724 $\pm$ 0.0077 \\
& LEARN-TS (Full)
& 0.5700 $\pm$ 0.0100
& 0.6025 $\pm$ 0.0094
& \textbf{0.5256 $\pm$ 0.0064}
& 0.7765 $\pm$ 0.0103 \\
\midrule

MSL
& w/o Obs.
& \textbf{0.3509 $\pm$ 0.0126}
& \textbf{0.3788 $\pm$ 0.0095}
& 0.3053 $\pm$ 0.0064
& 0.7576 $\pm$ 0.0103 \\
& Concat fusion
& \textbf{0.3509 $\pm$ 0.0065}
& 0.3733 $\pm$ 0.0074
& \textbf{0.3057 $\pm$ 0.0048}
& \textbf{0.7583 $\pm$ 0.0097} \\
& LEARN-TS (Full)
& 0.3472 $\pm$ 0.0184
& 0.3752 $\pm$ 0.0053
& 0.2951 $\pm$ 0.0023
& 0.7481 $\pm$ 0.0282 \\
\bottomrule
\end{tabular}
\end{table*}

% Uncomment after the dataset-wise results are finalized.
%
% \begin{table*}[t]
% \centering
% \caption{
% Dataset-wise ablation of cross-modal fusion.
% Results are reported as mean \(\pm\) standard deviation over five random
% seeds.
% The best result in each dataset--metric column is shown in \textbf{bold}.
% }
% \label{tab:datasetwise_cma_ablation}
% \footnotesize
% \setlength{\tabcolsep}{4pt}
% \renewcommand{\arraystretch}{1.1}
% \input{tables/datasetwise_cma_ablation}
% \end{table*}

% ---------------------------------------------------------
% Normality alignment and scoring
% ---------------------------------------------------------
\subsection{Normality Alignment and Scoring}
\label{app:datasetwise_normality_ablation}

The macro results in Table~\ref{tab:ablation_analysis}(b) are computed by
first averaging the four datasets equally within each seed and then
reporting the mean and sample standard deviation across the five seed-level
averages.
Table~\ref{tab:normality_ablation_datasetwise} reports the corresponding
dataset-wise results.

The ablation compares four configurations.
\textit{No normality} removes both alignment training and
discrepancy-guided scoring;
\textit{Scoring only} applies discrepancy-guided scoring without alignment;
\textit{Alignment only} retains alignment training but uses reconstruction
evidence alone at inference; and
\textit{LEARN-TS (Full)} uses both components.

The joint configuration substantially improves A-PR and V-PR on SWaT and
provides consistent ranking gains on SMD, whereas neither component alone
reproduces these improvements.
Its effects on PSM and MSL are limited or metric-dependent, and the
threshold-dependent metrics do not exhibit uniform gains.
These results indicate that normality guidance primarily benefits
anomaly-score ranking on datasets where reconstruction evidence alone
provides insufficient separation, rather than uniformly improving event
localization across all benchmarks.

\begin{table*}[t]
\caption{
Dataset-wise ablation of normality alignment and discrepancy-guided scoring.
Results are reported as mean \(\pm\) sample standard deviation over five
random seeds.
The best result in each dataset--metric column is shown in \textbf{bold}.
}
\label{tab:normality_ablation_datasetwise}
\centering
\footnotesize
\setlength{\tabcolsep}{4pt}
\renewcommand{\arraystretch}{1.1}

\begin{tabular*}{\textwidth}{
    @{\extracolsep{\fill}}
    lcccccc
    @{}
}
\toprule
Variant & Align. & Score & A-PR & V-PR & R-F1 & Aff-F1 \\
\midrule

\multicolumn{7}{l}{\textit{(a) SWaT}} \\
\addlinespace[2pt]

No normality
& \xmark & \xmark
& 0.2088 \(\pm\) 0.0076
& 0.1748 \(\pm\) 0.0010
& 0.2971 \(\pm\) 0.0611
& \textbf{0.7407 \(\pm\) 0.0157} \\

Scoring only
& \xmark & \cmark
& 0.1460 \(\pm\) 0.0531
& 0.1530 \(\pm\) 0.0264
& 0.2711 \(\pm\) 0.0635
& 0.7103 \(\pm\) 0.0271 \\

Alignment only
& \cmark & \xmark
& 0.2004 \(\pm\) 0.0094
& 0.1766 \(\pm\) 0.0008
& 0.2646 \(\pm\) 0.0211
& 0.7299 \(\pm\) 0.0097 \\

LEARN-TS (Full)
& \cmark & \cmark
& \textbf{0.7578 \(\pm\) 0.0117}
& \textbf{0.4859 \(\pm\) 0.0346}
& \textbf{0.3102 \(\pm\) 0.0436}
& 0.7277 \(\pm\) 0.0069 \\

\midrule
\multicolumn{7}{l}{\textit{(b) SMD}} \\
\addlinespace[2pt]

No normality
& \xmark & \xmark
& 0.4138 \(\pm\) 0.0043
& 0.4895 \(\pm\) 0.0037
& 0.3393 \(\pm\) 0.0041
& 0.8067 \(\pm\) 0.0078 \\

Scoring only
& \xmark & \cmark
& 0.4514 \(\pm\) 0.0222
& 0.5200 \(\pm\) 0.0158
& 0.3340 \(\pm\) 0.0033
& 0.8088 \(\pm\) 0.0129 \\

Alignment only
& \cmark & \xmark
& 0.4144 \(\pm\) 0.0031
& 0.4900 \(\pm\) 0.0003
& \textbf{0.3412 \(\pm\) 0.0019}
& 0.8099 \(\pm\) 0.0071 \\

LEARN-TS (Full)
& \cmark & \cmark
& \textbf{0.4930 \(\pm\) 0.0056}
& \textbf{0.5398 \(\pm\) 0.0037}
& 0.3203 \(\pm\) 0.0033
& \textbf{0.8121 \(\pm\) 0.0056} \\

\midrule
\multicolumn{7}{l}{\textit{(c) PSM}} \\
\addlinespace[2pt]

No normality
& \xmark & \xmark
& \textbf{0.5743 \(\pm\) 0.0091}
& 0.6056 \(\pm\) 0.0075
& 0.5194 \(\pm\) 0.0199
& 0.7759 \(\pm\) 0.0115 \\

Scoring only
& \xmark & \cmark
& 0.5741 \(\pm\) 0.0099
& \textbf{0.6083 \(\pm\) 0.0079}
& 0.5006 \(\pm\) 0.0132
& 0.7765 \(\pm\) 0.0129 \\

Alignment only
& \cmark & \xmark
& 0.5700 \(\pm\) 0.0101
& 0.6025 \(\pm\) 0.0094
& 0.5249 \(\pm\) 0.0119
& \textbf{0.7808 \(\pm\) 0.0079} \\

LEARN-TS (Full)
& \cmark & \cmark
& 0.5700 \(\pm\) 0.0100
& 0.6025 \(\pm\) 0.0094
& \textbf{0.5256 \(\pm\) 0.0064}
& 0.7765 \(\pm\) 0.0103 \\

\midrule
\multicolumn{7}{l}{\textit{(d) MSL}} \\
\addlinespace[2pt]

No normality
& \xmark & \xmark
& \textbf{0.3491 \(\pm\) 0.0122}
& 0.3733 \(\pm\) 0.0066
& 0.3058 \(\pm\) 0.0044
& 0.7612 \(\pm\) 0.0077 \\

Scoring only
& \xmark & \cmark
& 0.3489 \(\pm\) 0.0125
& 0.3732 \(\pm\) 0.0064
& \textbf{0.3066 \(\pm\) 0.0028}
& \textbf{0.7619 \(\pm\) 0.0077} \\

Alignment only
& \cmark & \xmark
& 0.3476 \(\pm\) 0.0162
& 0.3734 \(\pm\) 0.0045
& 0.3010 \(\pm\) 0.0061
& 0.7613 \(\pm\) 0.0087 \\

LEARN-TS (Full)
& \cmark & \cmark
& 0.3472 \(\pm\) 0.0184
& \textbf{0.3752 \(\pm\) 0.0053}
& 0.2951 \(\pm\) 0.0023
& 0.7481 \(\pm\) 0.0282 \\

\bottomrule
\end{tabular*}
\end{table*}

\begin{table}[!t]
\caption{
Dataset-wise comparison between random semantic-free and
dataset-agnostic semantic normality references.
Results are reported as mean \(\pm\) sample standard deviation over five
random seeds.
The best result in each dataset--metric column is shown in \textbf{bold};
values tied at the displayed precision are both highlighted.
}
\label{tab:datasetwise_reference_ablation}
\centering
\footnotesize
\setlength{\tabcolsep}{3pt}
\renewcommand{\arraystretch}{0.88}

\begin{tabular*}{\textwidth}{
    @{\extracolsep{\fill}}
    lcccc
    @{}
}
\toprule
Normality reference & A-PR & V-PR & R-F1 & Aff-F1 \\
\midrule

\multicolumn{5}{l}{\textit{(a) SWaT}} \\
\addlinespace[2pt]

Random semantic-free
& 0.7431 \(\pm\) 0.0426
& 0.4758 \(\pm\) 0.0364
& 0.2933 \(\pm\) 0.0182
& 0.7276 \(\pm\) 0.0017 \\

Dataset-agnostic semantic
& \textbf{0.7578 \(\pm\) 0.0117}
& \textbf{0.4859 \(\pm\) 0.0346}
& \textbf{0.3102 \(\pm\) 0.0436}
& \textbf{0.7277 \(\pm\) 0.0069} \\

\midrule
\multicolumn{5}{l}{\textit{(b) SMD}} \\
\addlinespace[2pt]

Random semantic-free
& 0.4919 \(\pm\) 0.0052
& 0.5388 \(\pm\) 0.0052
& 0.3153 \(\pm\) 0.0058
& 0.8022 \(\pm\) 0.0143 \\

Dataset-agnostic semantic
& \textbf{0.4930 \(\pm\) 0.0056}
& \textbf{0.5398 \(\pm\) 0.0037}
& \textbf{0.3203 \(\pm\) 0.0033}
& \textbf{0.8121 \(\pm\) 0.0056} \\

\midrule
\multicolumn{5}{l}{\textit{(c) PSM}} \\
\addlinespace[2pt]

Random semantic-free
& \textbf{0.5700 \(\pm\) 0.0089}
& \textbf{0.6031 \(\pm\) 0.0080}
& \textbf{0.5293 \(\pm\) 0.0102}
& 0.7764 \(\pm\) 0.0090 \\

Dataset-agnostic semantic
& \textbf{0.5700 \(\pm\) 0.0100}
& 0.6025 \(\pm\) 0.0094
& 0.5256 \(\pm\) 0.0064
& \textbf{0.7765 \(\pm\) 0.0103} \\

\midrule
\multicolumn{5}{l}{\textit{(d) MSL}} \\
\addlinespace[2pt]

Random semantic-free
& 0.3390 \(\pm\) 0.0117
& 0.3683 \(\pm\) 0.0028
& \textbf{0.3004 \(\pm\) 0.0055}
& 0.7429 \(\pm\) 0.0082 \\

Dataset-agnostic semantic
& \textbf{0.3472 \(\pm\) 0.0184}
& \textbf{0.3752 \(\pm\) 0.0053}
& 0.2951 \(\pm\) 0.0023
& \textbf{0.7481 \(\pm\) 0.0282} \\

\bottomrule
\end{tabular*}
\end{table}

% ---------------------------------------------------------
% Semantic-reference control
% ---------------------------------------------------------
\subsection{Normality-Reference Control}
\label{app:datasetwise_reference_control}

Table~\ref{tab:datasetwise_reference_ablation} examines the contribution
of semantic content in the normality reference. The random semantic-free
variant replaces the normality-prompt embedding with a fixed random
reference while preserving the alignment and scoring pipeline.

The semantic reference improves all four metrics on SWaT and SMD,
and A-PR, V-PR, and Aff-F1 on MSL, whereas the random reference
achieves higher R-F1 on MSL. The two variants perform comparably on PSM.

Together with the alignment-and-scoring ablation, these results
distinguish the benefits of the guidance mechanism from those of
reference content. Joint alignment and scoring yield substantial
ranking gains on SWaT and SMD, while semantic rather than random
references provide modest, dataset-dependent improvements.
These findings support the utility of semantic references without
establishing their necessity.

% \begin{table}[t]
% \centering
% \caption{LLM replacement control with a single fixed seed (42).
% MLP (w/o LLM) replaces both observation and normality embeddings.
% Macro averages weight the four datasets equally.
% Bold denotes the better result within each pair.}
% \label{tab:obs_encoder_control}
% \small
% \setlength{\tabcolsep}{6pt}
% \begin{tabular}{llcccc}
% \toprule
% Dataset & Variant & A-PR & V-PR & R-F1 & Aff-F1 \\
% \midrule
% SWaT
%  & GPT-2 (Full)
%  & \textbf{0.7684} & \textbf{0.5139}
%  & \textbf{0.3784} & 0.7343 \\
%  & MLP (w/o LLM)
%  & 0.7661 & 0.5058 & 0.2514 & \textbf{0.7353} \\
% \midrule
% SMD
%  & GPT-2 (Full)
%  & 0.4868 & \textbf{0.5376}
%  & \textbf{0.3226} & 0.8163 \\
%  & MLP (w/o LLM)
%  & \textbf{0.4876} & 0.5267 & 0.3164 & \textbf{0.8212} \\
% \midrule
% PSM
%  & GPT-2 (Full)
%  & \textbf{0.5793} & \textbf{0.6136}
%  & \textbf{0.5256} & 0.7659 \\
%  & MLP (w/o LLM)
%  & 0.5629 & 0.5944 & 0.4957 & \textbf{0.7798} \\
% \midrule
% MSL
%  & GPT-2 (Full)
%  & \textbf{0.3644} & \textbf{0.3815}
%  & 0.2957 & \textbf{0.7717} \\
%  & MLP (w/o LLM)
%  & 0.3543 & 0.3743 & \textbf{0.2985} & 0.7652 \\
% \midrule
% Macro
%  & GPT-2 (Full)
%  & \textbf{0.5498} & \textbf{0.5116}
%  & \textbf{0.3806} & 0.7721 \\
%  & MLP (w/o LLM)
%  & 0.5427 & 0.5003 & 0.3405 & \textbf{0.7754} \\
% \bottomrule
% \end{tabular}
% \end{table}

\begin{table*}[t]
\caption{
Dataset-wise sensitivity to
(a) the full-reconstruction weight \(\lambda_{\mathrm{full}}\) and
(b) the normality-alignment weight \(\lambda_{\mathrm{norm}}\).
Results are obtained using a fixed random seed.
Bold rows denote the configurations used in the main experiments.
}
\label{tab:datasetwise_loss_sensitivity}
\centering
\footnotesize
\renewcommand{\arraystretch}{0.90}
\setlength{\tabcolsep}{3pt}

\begin{tabular*}{0.86\textwidth}{
    @{\extracolsep{\fill}}
    llcccc
    @{}
}

% =========================================================
% (a) lambda_full
% =========================================================
\toprule
\multicolumn{6}{c}{
\textit{(a) Full-reconstruction weight
\(\lambda_{\mathrm{full}}\)}
} \\
\midrule
Dataset & Value & A-PR & V-PR & R-F1 & Aff-F1 \\
\midrule

SWaT & 0.0 & 0.1766 & 0.1738 & 0.2920 & 0.7381 \\
     & 0.1 & 0.7654 & 0.5004 & 0.2793 & 0.7121 \\
     & 0.2 & 0.7662 & 0.5122 & 0.3293 & 0.7332 \\
     & \textbf{0.5}
     & \textbf{0.7684}
     & \textbf{0.5139}
     & \textbf{0.3784}
     & \textbf{0.7343} \\
     & 1.0 & 0.7665 & 0.5023 & 0.2655 & 0.7276 \\
\cmidrule(lr){1-6}

SMD  & 0.0 & 0.4552 & 0.5277 & 0.3361 & 0.8128 \\
     & 0.1 & 0.4924 & 0.5394 & 0.3308 & 0.8122 \\
     & 0.2 & 0.4918 & 0.5383 & 0.3233 & 0.8151 \\
     & \textbf{0.5}
     & \textbf{0.4868}
     & \textbf{0.5376}
     & \textbf{0.3226}
     & \textbf{0.8163} \\
     & 1.0 & 0.4877 & 0.5367 & 0.3105 & 0.7976 \\
\cmidrule(lr){1-6}

PSM  & 0.0 & 0.5939 & 0.6198 & 0.4891 & 0.7631 \\
     & 0.1 & 0.5693 & 0.5989 & 0.5520 & 0.7873 \\
     & 0.2 & 0.5792 & 0.6106 & 0.5379 & 0.7628 \\
     & \textbf{0.5}
     & \textbf{0.5793}
     & \textbf{0.6136}
     & \textbf{0.5256}
     & \textbf{0.7659} \\
     & 1.0 & 0.5675 & 0.6031 & 0.5294 & 0.7716 \\
\cmidrule(lr){1-6}

MSL  & 0.0 & 0.3122 & 0.3549 & 0.3075 & 0.7566 \\
     & 0.1 & 0.3532 & 0.3765 & 0.3032 & 0.7580 \\
     & 0.2 & 0.3496 & 0.3734 & 0.2952 & 0.7654 \\
     & \textbf{0.5}
     & \textbf{0.3644}
     & \textbf{0.3815}
     & \textbf{0.2957}
     & \textbf{0.7717} \\
     & 1.0 & 0.3285 & 0.3736 & 0.2995 & 0.7692 \\

% =========================================================
% (b) lambda_norm
% =========================================================
\midrule
\multicolumn{6}{c}{
\textit{(b) Normality-alignment weight
\(\lambda_{\mathrm{norm}}\)}
} \\
\midrule
Dataset & Value & A-PR & V-PR & R-F1 & Aff-F1 \\
\midrule

SWaT & 0.000 & 0.1167 & 0.1502 & 0.3440 & 0.7298 \\
     & 0.005 & 0.7558 & 0.4773 & 0.2599 & 0.7303 \\
     & 0.010 & 0.7635 & 0.5025 & 0.2848 & 0.7114 \\
     & \textbf{0.020}
     & \textbf{0.7684}
     & \textbf{0.5139}
     & \textbf{0.3784}
     & \textbf{0.7343} \\
     & 0.050 & 0.7610 & 0.4963 & 0.2876 & 0.7486 \\
\cmidrule(lr){1-6}

SMD  & 0.000 & 0.4323 & 0.5044 & 0.3310 & 0.7877 \\
     & 0.005 & 0.4924 & 0.5345 & 0.3165 & 0.7926 \\
     & \textbf{0.010}
     & \textbf{0.4868}
     & \textbf{0.5376}
     & \textbf{0.3226}
     & \textbf{0.8163} \\
     & 0.020 & 0.4985 & 0.5402 & 0.3251 & 0.8179 \\
     & 0.050 & 0.4824 & 0.5373 & 0.3180 & 0.8103 \\
\cmidrule(lr){1-6}

PSM  & 0.000 & 0.5845 & 0.6173 & 0.4928 & 0.7742 \\
     & 0.005 & 0.5788 & 0.6134 & 0.5218 & 0.7667 \\
     & \textbf{0.010}
     & \textbf{0.5793}
     & \textbf{0.6136}
     & \textbf{0.5256}
     & \textbf{0.7659} \\
     & 0.020 & 0.5778 & 0.6127 & 0.5282 & 0.7758 \\
     & 0.050 & 0.5708 & 0.6050 & 0.5066 & 0.7873 \\
\cmidrule(lr){1-6}

MSL  & 0.000 & 0.3602 & 0.3800 & 0.3064 & 0.7662 \\
     & 0.005 & 0.3476 & 0.3790 & 0.3044 & 0.7668 \\
     & \textbf{0.010}
     & \textbf{0.3644}
     & \textbf{0.3815}
     & \textbf{0.2957}
     & \textbf{0.7717} \\
     & 0.020 & 0.3592 & 0.3801 & 0.3065 & 0.7705 \\
     & 0.050 & 0.3537 & 0.3810 & 0.2995 & 0.7559 \\

\bottomrule
\end{tabular*}
\end{table*}

% \subsection{LLM replacement control}
% \label{app:obs_encoder_control}
% The MLP (w/o LLM) variant replaces both language-derived embedding
% inputs with MLP outputs. The observation branch encodes the descriptors
% used in prompt construction, while the normality branch uses a fixed
% constant input shared across all windows. Both outputs match the
% corresponding GPT-2 embedding shapes and use the existing shared
% projection. Table~\ref{tab:obs_encoder_control} reports results
% using a single fixed seed (42).

% =========================================================
% Dataset-Wise Sensitivity Analysis
% =========================================================
\section{Dataset-Wise Sensitivity Analysis}
\label{app:datasetwise_sensitivity}

The sensitivity plots in the main paper present macro averaged trends across
the four datasets.
Tables~\ref{tab:datasetwise_loss_sensitivity} and
\ref{tab:datasetwise_scoring_masking_sensitivity} report the corresponding
dataset-wise results for the three weighting hyperparameters and the number
of masked training patches.
All sensitivity experiments use a single fixed random seed and are therefore
intended to characterize response trends rather than statistical variability.
For each weighting hyperparameter, the zero setting removes the corresponding
contribution while leaving the remaining architecture and evaluation pipeline
unchanged.
Bold rows indicate the configurations adopted in the main experiments for
the corresponding dataset.

\begin{table*}[t]
\caption{
Dataset-wise sensitivity to
(a) the discrepancy-gating coefficient \(\lambda_{\mathrm{gate}}\) and
(b) the number of masked training patches.
Results are obtained using a fixed random seed.
Bold rows denote the configurations used in the main experiments.
}
\label{tab:datasetwise_scoring_masking_sensitivity}
\centering
\footnotesize
\renewcommand{\arraystretch}{0.90}
\setlength{\tabcolsep}{3pt}

\begin{tabular*}{0.86\textwidth}{
    @{\extracolsep{\fill}}
    llcccc
    @{}
}

% =========================================================
% (a) lambda_gate
% =========================================================
\toprule
\multicolumn{6}{c}{
\textit{(a) Discrepancy-gating coefficient
\(\lambda_{\mathrm{gate}}\)}
} \\
\midrule
Dataset & Value & A-PR & V-PR & R-F1 & Aff-F1 \\
\midrule

SWaT & 0.00 & 0.2132 & 0.1763 & 0.2607 & 0.7387 \\
     & 0.01 & 0.7622 & 0.5137 & 0.2786 & 0.7330 \\
     & 0.05 & 0.7640 & 0.5138 & 0.2861 & 0.7338 \\
     & 0.10 & 0.7670 & 0.5153 & 0.3121 & 0.7337 \\
     & 0.50 & 0.7686 & 0.5138 & 0.3299 & 0.7394 \\
     & \textbf{1.00}
     & \textbf{0.7684}
     & \textbf{0.5139}
     & \textbf{0.3784}
     & \textbf{0.7343} \\
\cmidrule(lr){1-6}

SMD  & 0.00 & 0.4091 & 0.4897 & 0.3422 & 0.8107 \\
     & 0.01 & 0.4825 & 0.5399 & 0.3381 & 0.8154 \\
     & \textbf{0.05}
     & \textbf{0.4868}
     & \textbf{0.5376}
     & \textbf{0.3226}
     & \textbf{0.8163} \\
     & 0.10 & 0.4902 & 0.5369 & 0.3053 & 0.8134 \\
     & 0.50 & 0.4997 & 0.5290 & 0.2740 & 0.7811 \\
     & 1.00 & 0.5040 & 0.5231 & 0.2657 & 0.7673 \\
\cmidrule(lr){1-6}

PSM  & 0.00 & 0.5794 & 0.6136 & 0.5198 & 0.7742 \\
     & 0.01 & 0.5794 & 0.6136 & 0.5198 & 0.7743 \\
     & 0.05 & 0.5794 & 0.6136 & 0.5202 & 0.7749 \\
     & \textbf{0.10}
     & \textbf{0.5793}
     & \textbf{0.6136}
     & \textbf{0.5256}
     & \textbf{0.7659} \\
     & 0.50 & 0.5791 & 0.6136 & 0.5035 & 0.7713 \\
     & 1.00 & 0.5789 & 0.6137 & 0.5444 & 0.7628 \\
\cmidrule(lr){1-6}

MSL  & 0.00 & 0.3619 & 0.3772 & 0.3055 & 0.7705 \\
     & 0.01 & 0.3627 & 0.3782 & 0.3087 & 0.7709 \\
     & \textbf{0.05}
     & \textbf{0.3644}
     & \textbf{0.3815}
     & \textbf{0.2957}
     & \textbf{0.7717} \\
     & 0.10 & 0.3662 & 0.3820 & 0.3072 & 0.7759 \\
     & 0.50 & 0.3656 & 0.3807 & 0.2853 & 0.7601 \\
     & 1.00 & 0.3590 & 0.3695 & 0.2737 & 0.7201 \\

% =========================================================
% (b) number of masked patches
% =========================================================
\midrule
\multicolumn{6}{c}{
\textit{(b) Number of masked training patches}
} \\
\midrule
Dataset & \# Mask & A-PR & V-PR & R-F1 & Aff-F1 \\
\midrule

SWaT & \textbf{1}
     & \textbf{0.7684}
     & \textbf{0.5139}
     & \textbf{0.3784}
     & \textbf{0.7343} \\
     & 2 & 0.7423 & 0.4644 & 0.2901 & 0.7110 \\
     & 3 & 0.7456 & 0.4769 & 0.2673 & 0.7108 \\
\cmidrule(lr){1-6}

SMD  & \textbf{1}
     & \textbf{0.4868}
     & \textbf{0.5376}
     & \textbf{0.3226}
     & \textbf{0.8163} \\
     & 2 & 0.4901 & 0.5342 & 0.3147 & 0.8004 \\
     & 3 & 0.4998 & 0.5468 & 0.3205 & 0.8015 \\
\cmidrule(lr){1-6}

PSM  & \textbf{1}
     & \textbf{0.5793}
     & \textbf{0.6136}
     & \textbf{0.5256}
     & \textbf{0.7659} \\
     & 2 & 0.5687 & 0.6040 & 0.5158 & 0.7752 \\
     & 3 & 0.5703 & 0.6043 & 0.5188 & 0.7726 \\
\cmidrule(lr){1-6}

MSL  & \textbf{1}
     & \textbf{0.3644}
     & \textbf{0.3815}
     & \textbf{0.2957}
     & \textbf{0.7717} \\
     & 2 & 0.3656 & 0.3804 & 0.3040 & 0.7603 \\
     & 3 & 0.3626 & 0.3824 & 0.3079 & 0.7655 \\

\bottomrule
\end{tabular*}
\end{table*}

% ---------------------------------------------------------
% Loss and scoring weights
% ---------------------------------------------------------
\subsection{Loss and Scoring Weights}
\label{app:sens_weights}

Table~\ref{tab:datasetwise_loss_sensitivity} reports sensitivity to the
full-reconstruction weight \(\lambda_{\mathrm{full}}\) and the
normality-alignment weight \(\lambda_{\mathrm{norm}}\), while
Table~\ref{tab:datasetwise_scoring_masking_sensitivity}(a) reports
sensitivity to the discrepancy-gating coefficient
\(\lambda_{\mathrm{gate}}\).
The main experiments use \(\lambda_{\mathrm{full}}=0.5\) for all datasets,
\(\lambda_{\mathrm{norm}}=0.02\) for SWaT and \(0.01\) for SMD, PSM, and
MSL, and
\(\lambda_{\mathrm{gate}}\in\{1.0,0.05,0.1,0.05\}\) for SWaT, SMD, PSM,
and MSL, respectively.

SWaT is particularly sensitive to removing the full-window reconstruction
objective, normality alignment, or discrepancy-guided modulation, with
pronounced decreases in A-PR and V-PR when the corresponding weight is set
to zero.
Performance is generally more stable across positive weighting values,
although the preferred operating range remains dataset dependent.
PSM is comparatively insensitive to \(\lambda_{\mathrm{gate}}\) in A-PR
and V-PR, whereas SMD and MSL exhibit clearer trade-offs between ranking and
threshold-dependent event metrics as the gate strength increases.

% ---------------------------------------------------------
% Training masking
% ---------------------------------------------------------
\subsection{Training Masking}
\label{app:training_masking}

The main configuration masks one of the eight temporal patches during each
training forward pass.
Figure~\ref{fig:sensitivity}(d) presents the corresponding four-dataset
macro trend, while
Table~\ref{tab:datasetwise_scoring_masking_sensitivity}(b) provides the
complete dataset-wise results.
Masking additional patches generally degrades macro performance, with the
strongest effect observed on SWaT.
Although SMD and MSL exhibit small metric-dependent improvements under
heavier masking, these gains are not consistent across datasets or metrics.
Single-patch masking therefore provides the most balanced overall behavior
while matching the exhaustive single-patch masking scheme used at inference.

\section{Language-Backbone Robustness and Efficiency}
\label{app:backbone_efficiency}

We evaluate whether LEARN-TS remains effective across different frozen
language backbones while comparing their computational costs.
Table~\ref{tab:backbone_efficiency} reports efficiency and anomaly detection
performance across all four datasets for GPT-2,
Qwen2.5-1.5B~\citep{qwen2024qwen25}, and
Llama-3.2-1B~\citep{meta2024llama32}.

As shown in Table~\ref{tab:backbone_efficiency}, the three frozen
language backbones yield broadly comparable ranking performance
across all four datasets. GPT-2 achieves the highest embedding and
cached scoring throughput throughout, while detection performance
varies by dataset and metric. No backbone consistently outperforms
the others, and larger backbones do not yield uniform gains.

All efficiency measurements were performed on an NVIDIA L40S
GPU with 48 GB of GPU memory.
For SMD, detection metrics are macro-averaged across 28 machines
within each seed. Parameter counts report the range across machines,
peak training GPU memory reports the maximum, and scoring throughput
is computed as total windows divided by total scoring time.

\begin{table}[!t]
\caption{Backbone efficiency and detection performance on four
datasets. Detection results are mean $\pm$ sample standard
deviation over five seeds. Embedding throughput measures prompt
encoding; scoring throughput excludes text encoding and uses
cached embeddings. Bold indicates the best result within each
dataset and column.}
\label{tab:backbone_efficiency}
\centering
\renewcommand{\arraystretch}{1.1}

\textit{(a) Computational efficiency}
\vspace{0.8mm}

{\scriptsize
\begin{tabular*}{0.99\linewidth}{@{\extracolsep{\fill}}llcccc@{}}
\toprule
Dataset & Backbone & Trainable & Peak Train GPU & Embed. & Scoring \\
& & Params $\downarrow$ & (GiB) $\downarrow$
& (prompts/s) $\uparrow$ & (windows/s) $\uparrow$ \\
\midrule
SWaT & GPT-2 & \textbf{41.16M} & \textbf{0.795} & \textbf{80.90} & \textbf{1111.46} \\
& Qwen2.5-1.5B & 41.75M & 0.811 & 38.81 & 932.19 \\
& Llama-3.2-1B & 42.14M & 0.822 & 53.18 & 958.34 \\
\midrule
PSM & GPT-2 & \textbf{40.79M} & \textbf{0.792} & \textbf{80.15} & \textbf{1194.91} \\
& Qwen2.5-1.5B & 41.38M & 0.803 & 40.17 & 1152.70 \\
& Llama-3.2-1B & 41.77M & 0.814 & 55.89 & 1097.99 \\
\midrule
MSL & GPT-2 & \textbf{40.96M} & \textbf{0.794} & \textbf{80.12} & \textbf{1212.53} \\
& Qwen2.5-1.5B & 41.55M & 0.805 & 40.33 & 1168.39 \\
& Llama-3.2-1B & 41.95M & 0.815 & 56.41 & 1116.43 \\
\midrule
SMD & GPT-2 & \textbf{40.84--41.06M} & \textbf{0.794} & \textbf{82.11} & \textbf{1185.91} \\
& Qwen2.5-1.5B & 41.43--41.65M & 0.805 & 40.29 & 1098.63 \\
& Llama-3.2-1B & 41.82--42.04M & 0.816 & 56.14 & 1108.09 \\
\bottomrule
\end{tabular*}
}
\vspace{2mm}
\textit{(b) Anomaly detection performance}
\vspace{0.8mm}

{\scriptsize
\begin{tabular*}{0.99\linewidth}{@{\extracolsep{\fill}}llcccc@{}}
\toprule
Dataset & Backbone & A-PR & V-PR & R-F1 & Aff-F1 \\
\midrule
SWaT & GPT-2 & 0.7578 $\pm$ 0.0117 & 0.4859 $\pm$ 0.0346 & \textbf{0.3102 $\pm$ 0.0436} & \textbf{0.7277 $\pm$ 0.0069} \\
& Qwen2.5-1.5B & 0.7540 $\pm$ 0.0376 & 0.4883 $\pm$ 0.0320 & 0.2677 $\pm$ 0.0237 & 0.7204 $\pm$ 0.0078 \\
& Llama-3.2-1B & \textbf{0.7654 $\pm$ 0.0050} & \textbf{0.4955 $\pm$ 0.0117} & 0.2626 $\pm$ 0.0191 & 0.7199 $\pm$ 0.0166 \\
\midrule
PSM & GPT-2 & 0.5700 $\pm$ 0.0100 & 0.6025 $\pm$ 0.0094 & 0.5256 $\pm$ 0.0064 & 0.7765 $\pm$ 0.0103 \\
& Qwen2.5-1.5B & \textbf{0.5748 $\pm$ 0.0088} & \textbf{0.6029 $\pm$ 0.0072} & 0.5136 $\pm$ 0.0252 & 0.7783 $\pm$ 0.0145 \\
& Llama-3.2-1B & 0.5716 $\pm$ 0.0118 & 0.6027 $\pm$ 0.0096 & \textbf{0.5266 $\pm$ 0.0096} & \textbf{0.7787 $\pm$ 0.0056} \\
\midrule
MSL & GPT-2 & \textbf{0.3472 $\pm$ 0.0184} & \textbf{0.3752 $\pm$ 0.0053} & 0.2951 $\pm$ 0.0023 & 0.7481 $\pm$ 0.0282 \\
& Qwen2.5-1.5B & 0.3421 $\pm$ 0.0077 & 0.3717 $\pm$ 0.0107 & 0.3005 $\pm$ 0.0035 & 0.7552 $\pm$ 0.0246 \\
& Llama-3.2-1B & 0.3361 $\pm$ 0.0075 & 0.3688 $\pm$ 0.0046 & \textbf{0.3006 $\pm$ 0.0057} & \textbf{0.7620 $\pm$ 0.0045} \\
\midrule
SMD & GPT-2 & \textbf{0.4930 $\pm$ 0.0056} & \textbf{0.5398 $\pm$ 0.0037} & \textbf{0.3203 $\pm$ 0.0033} & 0.8121 $\pm$ 0.0056 \\
& Qwen2.5-1.5B & 0.4874 $\pm$ 0.0121 & 0.5342 $\pm$ 0.0030 & 0.3173 $\pm$ 0.0030 & \textbf{0.8125 $\pm$ 0.0058} \\
& Llama-3.2-1B & 0.4897 $\pm$ 0.0070 & 0.5341 $\pm$ 0.0047 & 0.3150 $\pm$ 0.0069 & 0.8105 $\pm$ 0.0123 \\
\bottomrule
\end{tabular*}
}
\end{table}

% =========================================================
% F. Prompt Details
% =========================================================
\section{Prompt Templates}
\label{app:prompt_templates}

LEARN-TS uses a window-specific observation prompt and a fixed,
dataset-agnostic normality prompt with distinct semantic roles.
Both are encoded by the same frozen language model and use neither anomaly
labels nor test-set-wide statistics.
Figure~\ref{fig:normality_prompt} shows the fixed normality prompt, while Figure~\ref{fig:swat_observation_prompt}
illustrates an abridged observation template with placeholders
for window-dependent fields.
LEARN-TS extracts token-level hidden states rather than generating or parsing
textual responses.

% ---------------------------------------------------------
% F.1 Normality Reference Construction
% ---------------------------------------------------------
\subsection{Normality Reference Construction}
\label{app:normality_references}

LEARN-TS uses the dataset-agnostic semantic reference as its default
normality reference.
To determine whether normality guidance benefits from semantic content rather
than merely from an additional reference input, we compare it with a random
semantic-free reference of the same input shape.

% ---------------------------------------------------------
% F.1.1 Dataset-Agnostic Semantic Reference
% ---------------------------------------------------------
\subsubsection{Dataset-Agnostic Semantic Reference}
\label{app:generic_normality_prompt}

The normality prompt is fixed before training and shared unchanged across
SWaT, SMD, PSM, and MSL.
It describes broad properties of normal multivariate temporal behavior
without using dataset-specific sensor identities, operating rules, anomaly
labels, or window-specific observations.

\begin{figure}[t]
    \centering
    \includegraphics[width=\linewidth]{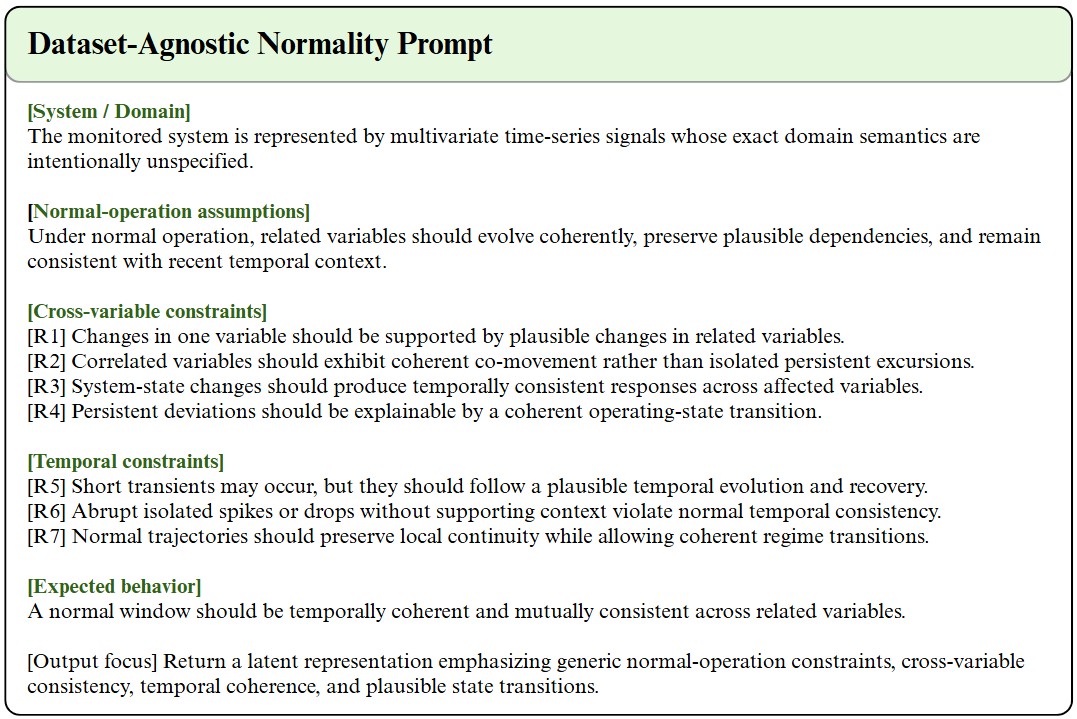}
    \caption{
    Dataset-agnostic normality prompt used in LEARN-TS.
    The prompt is fixed before training and shared unchanged across all
    datasets and input windows.
    }
    \label{fig:normality_prompt}
\end{figure}

Figure~\ref{fig:normality_prompt} shows the complete prompt used to construct
the shared normality reference.
The prompt encodes generic assumptions about cross variable consistency,
temporal coherence, plausible state transitions, and context-supported
recovery.
Because the same prompt is used for every dataset and input window, its
language model representation serves as a window-independent semantic
reference rather than as context associated with a particular observation.

\subsubsection{Random Semantic-Free Reference}
\label{app:random_normality_reference}

The random semantic-free reference evaluates whether normality guidance
benefits from semantic normality content rather than merely from an
additional reference input.
For each experimental run, we generate a seeded random tensor
\(\mathbf{E}_{\mathrm{rand}}\in\mathbb{R}^{T_n\times d_e}\) with the same
shape as the frozen normality-prompt embedding
\(\mathbf{E}_{\mathrm{norm}}
=E_{\mathrm{text}}(q^{\mathrm{norm}})\).
The random tensor replaces the frozen language model embedding before the
shared text projection:
\begin{equation}
    \mathbf{H}_{\mathrm{rand}}
    =
    \phi(\mathbf{E}_{\mathrm{rand}})
    \in\mathbb{R}^{T_n\times d}.
    \label{eq:random_normality_reference}
\end{equation}
The random input is sampled once for each experimental seed and remains fixed
throughout training and inference.
The shared projection and all downstream normality-alignment and
discrepancy-scoring components are otherwise identical to the dataset-agnostic semantic-reference configuration.
This control therefore preserves the input shape and downstream architecture
while removing language-derived normality content.

\begin{figure}[t]
    \centering
    \includegraphics[width=\linewidth]{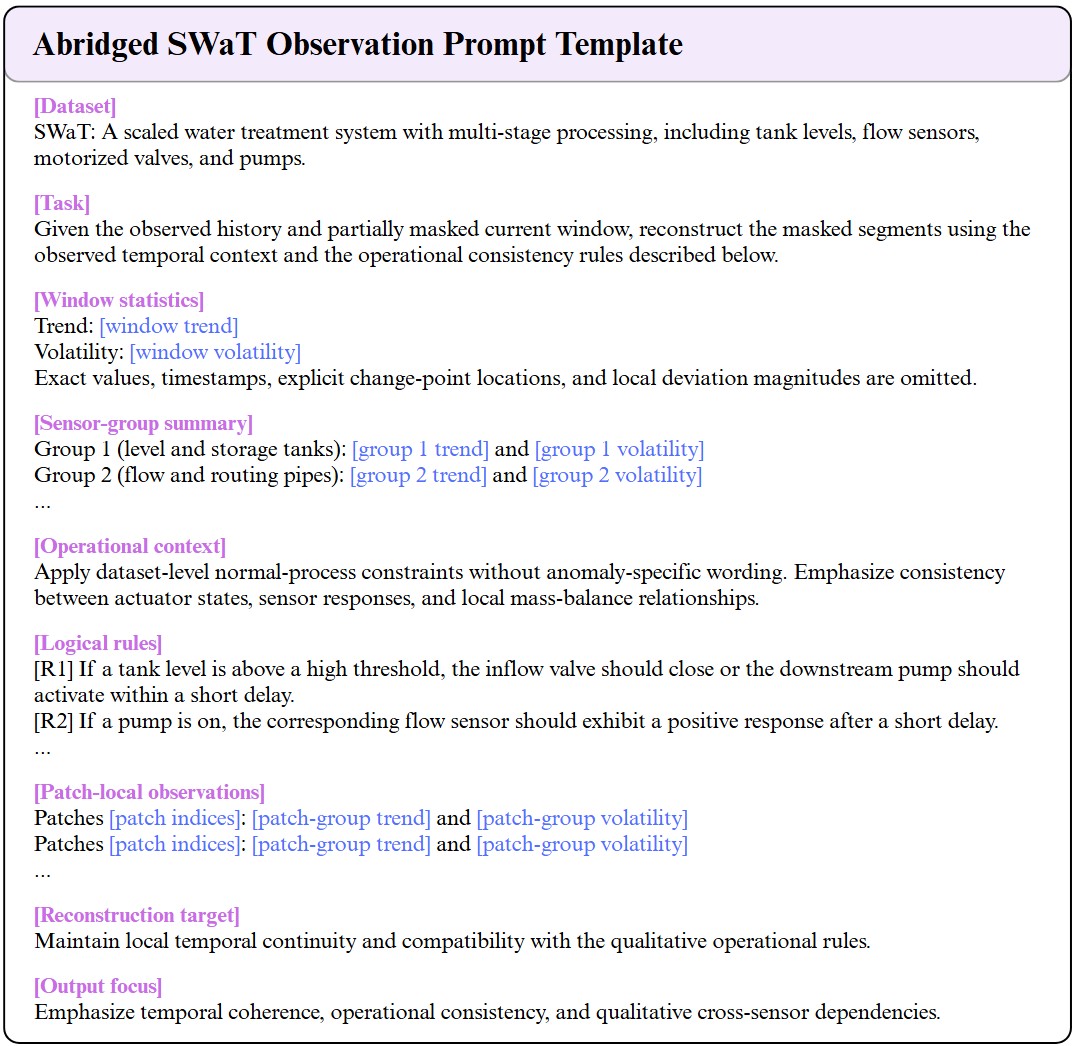}
\caption{
Abridged SWaT observation prompt template.
Blue placeholders denote window-dependent descriptors and
patch-index lists; dataset-level descriptions and rules remain fixed.
Exact measurements and explicit anomaly labels are excluded.
Repeated entries are omitted for readability.
}
    \label{fig:swat_observation_prompt}
\end{figure}

\subsection{Window-Specific Observation Prompt}
\label{app:observation_prompt}

The observation prompt combines a common template structure with
fixed dataset-specific profiles and deterministic window descriptors.
Within each dataset, the same profile and descriptor-construction
procedure are used for training, validation, and test windows.
Unlike the dataset-agnostic normality prompt, observation prompts
include dataset-level system descriptions, sensor-group definitions,
and qualitative operational rules alongside window-derived descriptors.
They require no temporally paired external text and use no anomaly labels.
Exact numerical measurements, timestamps, explicit change-point
locations, and local deviation magnitudes are excluded.
Figure~\ref{fig:swat_observation_prompt} shows an abridged SWaT
observation prompt template, retaining all information types
used by the implementation.
The same prompt structure is used for the other datasets, with their
corresponding system descriptions, sensor groups, operational rules, and
window-derived descriptors.

The descriptors are computed from the fully observed window before numerical
masking.
Consequently, they may retain coarse qualitative information about a
subsequently masked patch, such as its trend or volatility, without exposing
its exact numerical target.
The masking objective therefore suppresses direct numerical identity mapping
rather than defining information-free missing-value imputation.

\paragraph{Prompt structure and construction.}
Each prompt contains:
(i) a dataset-level system description and reconstruction task;
(ii) global trend and volatility descriptors;
(iii) sensor-group-level temporal summaries;
(iv) qualitative operational context and logical rules;
(v) patch-local trend and variability descriptors; and
(vi) the reconstruction objective.

All descriptors are computed deterministically using statistics from the
current window.
Volatility levels are assigned using fixed normalized cutoffs of \(0.3\) and
\(0.8\), and trend categories follow the deterministic rule implemented in
the released code.
Dataset-level descriptions are based on the original dataset papers
\citep{goh2017swat,su2019omnianomaly,abdulaal2021psm,hundman2018msl}.
Where available, accompanying public sensor or channel metadata are
additionally used to define channel-group keywords.
For SWaT, qualitative operational rules are manually specified based on
public descriptions of the testbed and its sensor metadata, covering coarse
relationships among tank levels, valve and pump states, flow responses, and
mass-balance consistency.
Channel groups are constructed by predefined keyword matching, with unmatched
channels assigned to auxiliary groups in their original order.
All dataset profiles are specified independently of anomaly labels and
test-set events and remain fixed across windows.
Complete keyword lists, operational rules, descriptor formulas, and
exception-handling procedures are provided in the released code.

\paragraph{Text encoding.}
The prompt is encoded once by the frozen language model, and its final-layer
token hidden states are cached for downstream use.
Padding tokens are excluded from representation pooling.
The tokenizer configuration and input-length handling are provided in the
released code.

\end{document}